\pdfoutput=1
\documentclass[11pt]{article}

\usepackage[preprint]{acl}

\usepackage{times}
\usepackage{latexsym}

\usepackage[T1]{fontenc}
\usepackage[utf8]{inputenc}

\usepackage{microtype}

\usepackage{microtype}
\usepackage{booktabs}       % professional-quality tables
\usepackage{colortbl}
\usepackage{nicefrac}       % compact symbols for 1/2, etc.
\usepackage{hyperref}
\usepackage{url}
\usepackage[oldenum]{paralist}
\usepackage{amsmath}
\usepackage{amsfonts}
\usepackage{amssymb,stmaryrd}
\usepackage{verbatim}
\usepackage{subfigure}
\usepackage[linesnumbered,lined,boxed,commentsnumbered,ruled,vlined]{algorithm2e}
\usepackage{algorithmic}
\usepackage{tikz}
\usepackage{pgfplots}
\pgfplotsset{compat=1.18} 
\usepackage{array}
\usepackage{multirow}
\usepackage{makecell}
\usepackage{enumerate}
\usepackage{enumitem}
\usepackage{bm}
\usepackage{comment}
\usepackage{soul}
\usepackage{footmisc}
\usepackage[normalem]{ulem}
\usepackage{minibox}
\usepackage{framed}
\graphicspath{{figures/}}
\usetikzlibrary{positioning}
\usepackage{pgfgantt}
\allowdisplaybreaks           		
\usepackage{graphicx}				
\usepackage{amssymb}
\usepackage{arydshln}
\usepackage{bm}
\usepackage{tcolorbox}
\usepackage{awesomebox}
\usepackage{pifont}
\usepackage[flushleft]{threeparttable} % add notes to the table 
\usepackage{lipsum}
\usepackage{inconsolata}

\newcommand{\jy}[1]{\textcolor{black}{#1}}

\def \OURS {KiRAG}

\definecolor{myblue}{HTML}{99ccff}
\definecolor{myorange}{HTML}{fde6cc}

\title{\OURS{}: Knowledge-Driven Iterative Retriever for Enhancing Retrieval-Augmented Generation}

\author{
Jinyuan Fang \\ University of Glasgow \\ \texttt{\small j.fang.2@research.gla.ac.uk} \\ \And
Zaiqiao Meng\thanks{Corresponding Author.} \\ University of Glasgow \\ \texttt{\small zaiqiao.meng@glasgow.ac.uk} \\ \And
Craig Macdonald \\ University of Glasgow \\ \texttt{\small craig.macdonald@glasgow.ac.uk} \\
}
\begin{document}
\maketitle
\begin{abstract}
Iterative \hspace{-0.4pt}retrieval-augmented \hspace{-0.4pt}generation \hspace{-1.2pt}({i}RAG) models offer an effective approach for multi-hop question answering (QA). However, their \textit{retrieval process} faces two key challenges: (1) it can be disrupted by {irrelevant} documents or factually inaccurate chain-of-thoughts; (2) their retrievers are not designed to dynamically adapt to the evolving information needs in multi-step reasoning, making it difficult to identify and retrieve the missing information required at each iterative step. Therefore, we propose \textbf{\OURS{}}\footnote{Code: \url{https://github.com/jyfang6/kirag}}, {which uses} a knowledge-driven iterative retriever model {to enhance} the retrieval process {of iRAG}. Specifically, \OURS{} decomposes documents into knowledge triples and performs iterative retrieval with these triples to {enable} a factually reliable retrieval process. Moreover, \OURS{} integrates reasoning into the retrieval process to dynamically identify and retrieve knowledge that bridges information gaps, effectively adapting to the evolving information needs. Empirical results show that \OURS{} {significantly} outperforms existing iRAG models, with an average improvement of 9.40\% in R$\text{@}3$ and 5.14\% in F1 on multi-hop QA. 
\end{abstract}

\section{Introduction}
Retrieval-augmented generation (RAG) models have demonstrated superior performance in question answering (QA) tasks~\cite{lewis2020retrieval,ram2023context,lin2024ra}. While standard RAG models excel at single-hop questions, they often struggle with multi-hop questions~\cite{trivedi2023interleaving}, which require reasoning over multiple interconnected pieces of information to derive correct answers. The key limitation is that their single-step retrieval process often fails to retrieve all the relevant information needed to answer multi-hop questions~\cite{shao2023enhancing}, leading to knowledge gaps in the reasoning process. To address this limitation, iterative RAG (\jy{iRAG)} models have been proposed~\cite{trivedi2023interleaving,asai2024self,su2024dragin,yao2024seakr}. These models employ multiple steps of retrieval and reasoning to iteratively gather the necessary information for addressing multi-hop questions.

\begin{table}[tb]
\centering
\resizebox{0.48\textwidth}{!}{
\begin{tabular}{l|l}
\hline \rule{0pt}{1.0em}

\multirow{2}{*}{\textbf{Question}}
& According to the 2001 census, what was the population \\
& of the city in which Kirton End is located? \\
\hline \rule{0pt}{1.0em}
\multirow{5}{*}{\textbf{\OURS{} (Ours)}}
& \textbf{Step 1}: \colorbox{myblue}{Kirton End} is a hamlet in the civil parish of \\ 
& Kirton in the \colorbox{myblue}{Boston district of Lincolnshire, England.}... \\
& \textbf{Step 2}: \colorbox{myblue}{Boston} is a town and small port in Lincolnshire, \\
& on the east coast of England... while the town itself \colorbox{myblue}{had a} \\
& \colorbox{myblue}{population of 35,124 at the 2001 census.}\\
\hline \rule{0pt}{1.0em}

\multirow{5}{*}{\textbf{IRCoT}}
& \textbf{Step 1}: \colorbox{myblue}{Kirton End} is a hamlet in the civil parish of \\ 
& Kirton in the \colorbox{myblue}{Boston district of Lincolnshire, England.}... \\ 
&  \textbf{Step 2}: \colorbox{myorange}{Kirton} is a village in Nottinghamshire, England... \\
& \colorbox{myorange}{According to the United Kingdom Census 2001 it had a} \\
& \colorbox{myorange}{population of 273}, reducing to 261 at the 2011 census. \\
\hline \rule{0pt}{1.0em}

\multirow{5}{*}{\textbf{IRDoc}}
& \textbf{Step 1}: \colorbox{myblue}{Kirton End} is a hamlet in the civil parish of \\ 
& Kirton in the \colorbox{myblue}{Boston district of Lincolnshire, England.}... \\ 
& \textbf{Step 2}: \colorbox{myorange}{Ollerton} is a small town in Nottinghamshire... \\ 
& \colorbox{myorange}{The population of this parish at the 2011 census was 9,840}.\\
\hline    
\end{tabular}
}
\label{table:motivation}
\end{table}
\begin{figure}[tb]
\vspace{-1em}
\begin{center}
\includegraphics[width=0.45\textwidth]{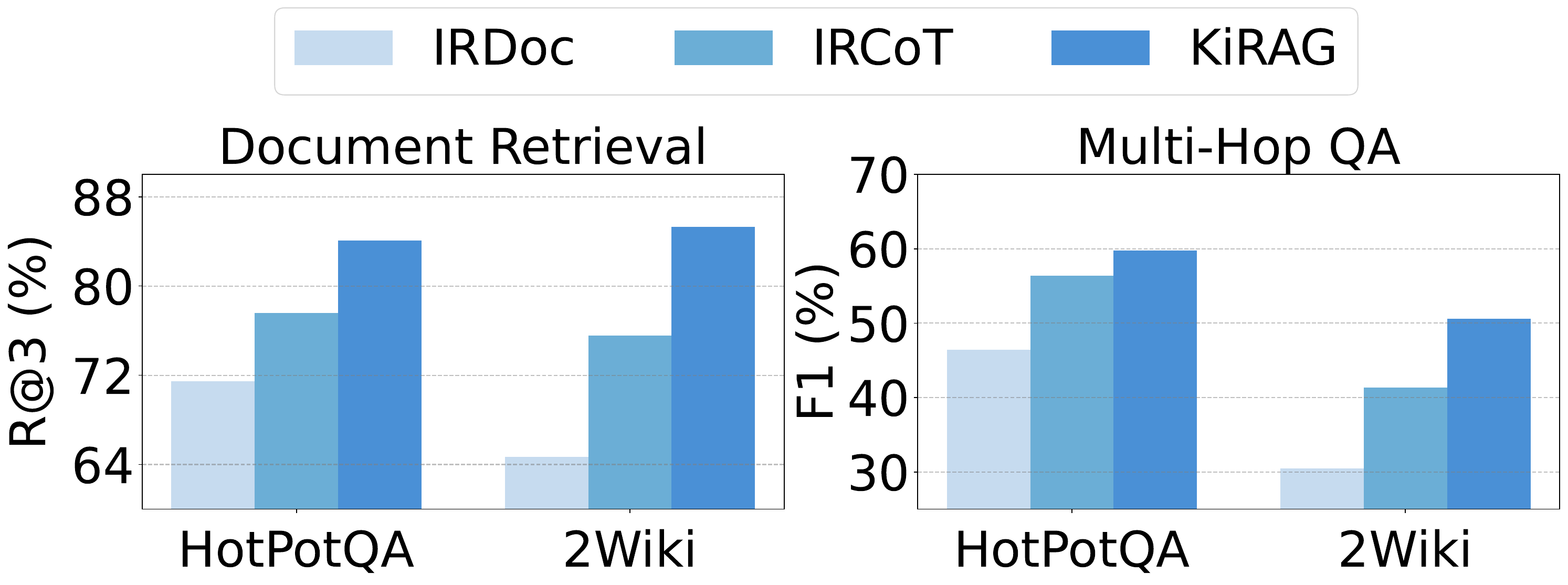}
\end{center}
\vspace{-0.8em}
\caption{
\textit{
(top)} Example of top-ranked documents at each step, with relevant content marked in \colorbox{myblue}{blue} and distracting content in \colorbox{myorange}{orange}. 
\jy{We compare \OURS{} with IRCoT~\cite{trivedi2023interleaving} and its variant IRDoc, where we replace generated thoughts with top-ranked documents.} 
\textit{(Bottom)} The corresponding retrieval and QA performance on HotPotQA and 2Wiki datasets. 
}
\label{figure:motivation}
\vspace{-1.0em}
\end{figure}

Despite the effectiveness of existing \jy{iRAG} models, their \textit{retrieval process} faces two key challenges:
(1) \jy{These models perform iterative retrieval by iteratively augmenting the query with either previously retrieved documents~\cite{zhao2021multi} or generated chain-of-thoughts~\cite{trivedi2023interleaving}.} 
However, retrieved documents often include noise or irrelevant information~\cite{yoran2024making}, while generated chain-of-thoughts can contain factually inaccurate content~\cite{wang2023survey,luo2024reasoning}. The propagation of these distracting contexts can degrade retrieval quality and ultimately hinder overall RAG performance.
(2) Answering a multi-hop question requires multi-step reasoning, where the information needed to derive the correct answer evolves with each iteration. For example, to answer the question in Figure~\ref{figure:motivation}, the first iterative step requires retrieving the location of Kirton End (Boston). Once this information is obtained, the next step shifts to retrieving Boston's population in 2001, demonstrating how the information needed to answer a multi-hop question evolves with each iteration. However, existing iRAG models often rely on off-the-shelf retrieval models that retrieve information based on semantic similarity. These retrievers are not designed to dynamically adapt to the evolving information needs in multi-step reasoning, making it difficult to identify and retrieve the missing pieces of information needed at each iteration, thereby hindering the overall retrieval effectiveness. Figure~\ref{figure:motivation} illustrates these two key challenges, highlighting the necessity of developing a retrieval approach that can mitigate the impact of irrelevant documents or inaccurate thoughts, and dynamically adapt to evolving information needs.

{To this end, we propose} \textbf{\OURS{}}, \jy{which leverages} a \textbf{K}nowledge-driven iterative retriever model \jy{to enhance} the retrieval process of \textbf{iRAG} models. Specifically, to address the challenge of \jy{irrelevant} documents and inaccurate thoughts, \jy{inspired by prior works~\cite{fang2024reano,fang2024trace} that use knowledge triples for enhanced reasoning}, \OURS{} decomposes documents into knowledge triples, formatted as $\langle$\textit{head entity}, \textit{relation}, \textit{tail entity}$\rangle$, and performs iterative retrieval with these triples. By leveraging knowledge triples, which are compact and grounded in documents, \OURS{} \jy{enables} a more focused and factually reliable retrieval process. 

Moreover, \jy{to address the challenge of evolving information needs}, \OURS{} employs a knowledge-driven iterative retrieval framework to retrieve relevant knowledge triples from the corpus systematically. This framework \textit{integrates reasoning into retrieval process}, enabling the system to identify and retrieve knowledge that bridges information gaps dynamically. 
Specifically, the iterative retrieval process incrementally builds a \textit{knowledge triple-based reasoning chain}, such as ``$\langle$\textit{Kirton End; location; Boston}$\rangle$,$\langle$\textit{Boston}; \textit{population in 2001 census}; \textit{35,124}$\rangle$'', by retrieving triples step-by-step.  
At each iteration, given the current step reasoning chain, e.g., ``$\langle$\textit{Kirton End; location; Boston}$\rangle$'', \OURS{} dynamically identifies and retrieves the missing knowledge triples needed to coherently extend the chain towards answering the question. This targeted approach can effectively guide the retrieval process in acquiring multiple interconnected pieces of information needed for addressing \jy{a} question. 

We evaluate \OURS{} on five multi-hop and one single-hop QA datasets. \OURS{} outperforms existing iRAG models, achieving average improvements of 9.40\% in R$\text{@}$3 and 7.59\% in R$\text{@}$5 on multi-hop QA, which lead to an improvement of 5.14\% in F1. \jy{Despite that} \OURS{} is designed for multi-hop QA, it achieves comparable retrieval and QA performance with state-of-the-art baseline on the single-hop QA dataset, demonstrating its effectiveness across different types of questions. 

Our contributions can be summarised as follows:
(1) We propose \OURS{}, which performs iterative retrieval with knowledge triples to enhance the retrieval process of \jy{iRAG} models; (2) \OURS{} uses a knowledge-driven iterative retrieval framework to \jy{dynamically adapt the retrieval process to the evolving information needs in multi-step reasoning;} (3) Empirical results show that \OURS{} achieves superior performance on multi-hop QA. 

\section{Problem Formulation}
Our approach builds on the \jy{iRAG} process. \jy{Given a question $q$ and its answer $a$, iRAG} is formalised as:
\begin{align}
    & \scalebox{0.99}{$p_{\theta,\phi}(a | q, \mathcal{C}) \sim p_\phi(a | q, \mathcal{D}_q) p_\theta( \mathcal{D}_q | q, \mathcal{C})$}, \\
    & \scalebox{0.99}{$p_\theta( \mathcal{D}_q | q, \mathcal{C}) \sim \prod_{i=1}^L p_\theta(\mathcal{D}_q^i | q, \mathcal{D}_q^{<i})$},
\end{align}
where $p_\theta$ denotes the \textit{retriever model} \jy{that iteratively retrieves documents $\mathcal{D}_q$=$\{\mathcal{D}_q^i\}_{i=1}^L$ from a corpus $\mathcal{C}$} and $p_\phi$ is the \textit{reader model}.  
\jy{At the $i$-th iteration, the retriever model retrieves documents $\mathcal{D}_q^i$ based on question $q$ and previously retrieved documents $\mathcal{D}_q^{<i}$.} 
In this paper, we primarily focus on enhancing the \textit{retriever model}, $p_\theta$, to effectively retrieve relevant documents from the corpus. 
To evaluate the effectiveness of our approach, we focus on multi-hop QA, a standard \jy{type of} benchmark for assessing \jy{iRAG} systems~\cite{gao2023retrieval}. 

\begin{figure*}[!t]
\vspace{-0.9em}
\begin{center}
\includegraphics[width=0.9\textwidth]{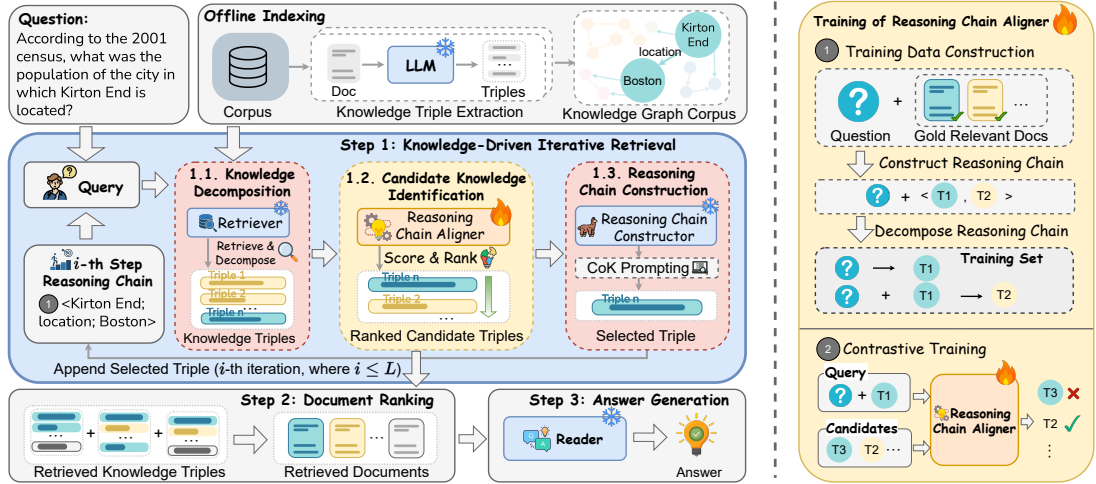}
\end{center}
\vspace{-0.8em}
\caption{
\textit{(left)} Overview of \OURS{}. Given a question, it employs a knowledge-driven iterative retrieval process (\textbf{Step 1}) to retrieve relevant knowledge triples, including three iterative steps: knowledge decomposition, candidate knowledge identification and reasoning chain construction. The retrieved triples are used to rank documents (\textbf{Step 2}), which are passed to the reader for answer generation (\textbf{Step 3}).
\textit{(right)} Training strategy for the Reasoning Chain Aligner, designed to optimise the identification of relevant knowledge triples at each step of the retrieval process.
}
\label{figure:model}
\vspace{-0.5em}
\end{figure*}

\section{\OURS{}}
This section begins with an overview of \OURS{} in \S\ref{subsec:overview}. 
Next, we present a detailed explanation of each component from \S\ref{subsec:iterative_retrieval} to \S\ref{subsec:document_ranking}. Finally, the training strategy is introduced in \S\ref{subsec:training}. 

\subsection{Overview}
\label{subsec:overview}
Figure~\ref{figure:model} provides an overview of our approach. \OURS{} uses a \textit{knowledge-driven iterative retrieval} framework to systematically retrieve a comprehensive set of relevant knowledge triples $\mathcal{T}_q$ (see \S\ref{subsec:iterative_retrieval}). \jy{Next, it leverages the retrieved knowledge triples $\mathcal{T}_q$ to identify and rank documents based on their relevance to the question (see \S\ref{subsec:document_ranking})}. Therefore, \jy{the retriever model} of \OURS{} can be formulated as: 
\begin{align}
\label{eq:kirag_retriever_model}
    \scalebox{0.85}{$p_\theta(\mathcal{D}_q | q, \mathcal{C}) \sim p_\theta(\mathcal{D}_q|q, \mathcal{T}_q) \prod_{i=1}^L p_\theta(\mathcal{T}_{q}^i | q, \mathcal{T}_{q}^{<i} , \mathcal{C})$}, 
\end{align}
where $\mathcal{T}_{q}^i$ is the set of knowledge triples retrieved at the $i$-th iteration, $\mathcal{T}_{q}^{<i}$ represents all previously retrieved triples and $L$ is the maximum number of iterations. Once we obtain the retrieved documents $\mathcal{D}_q$, \OURS{} employs an LLM-based reader model $q_\phi$ to generate the answer to the question. 

\subsection{\jy{Knowledge-Driven Iterative Retrieval}}
\label{subsec:iterative_retrieval} 
\OURS{} retrieves relevant knowledge triples from the corpus by progressively building a \textit{knowledge triple-based reasoning chain}, i.e., a sequence of logically connected knowledge triples that support answering a given question. For instance, the chain \textit{$\langle$\textit{Kirton End; location; Boston}$\rangle$,$\langle$\textit{Boston}; \textit{population in 2001 census}; \textit{35,124}$\rangle$} provides relevant knowledge for answering the question in Figure~\ref{figure:model}. The reasoning chain is built iteratively by selecting triples step-by-step. At the $i$-th iteration, given the \textit{$i$-th step reasoning chain}, which is a sequence of triples obtained up to the $i$-th iteration, such as \textit{$\langle$\textit{Kirton End; location; Boston}$\rangle$}, the framework retrieves and selects the next triple to extend the reasoning chain through the following three steps:

\vspace{0.5em} \noindent \textbf{Knowledge Decomposition}. 
\jy{To enable a factually reliable retrieval process,} \OURS{} decomposes documents into knowledge triples. 
\jy{At the $i$-th iteration, the query $q_i$ is formed} by concatenating the question with the $i$-th step reasoning chain in the format ``\textit{\{question\}. knowledge triples: \{triple1\}...}''. \OURS{} employs an off-the-shelf \textit{Retriever} model to retrieve \jy{$K_0$}\footnote{We provide analysis of the effect of $K_0$ in Appendix~\ref{app:effect_of_m}.} documents from the corpus, providing an initial pool of information for extracting relevant knowledge \jy{(see Step 1.1 in Figure~\ref{figure:model})}. 

Building on recent advancements in extracting knowledge triples using LLMs~\cite{edge2024local,fang2024trace}, we employ in-context learning to prompt an LLM to extract knowledge triples for each retrieved document independently. Since the extraction process is query-independent, triples can be precomputed offline for all documents in the corpus\footnote{The knowledge triples for retrieved documents can be obtained using the document IDs during the retrieval process.}. This enables the construction of a \textit{knowledge graph (KG) corpus}, effectively improving retrieval efficiency\footnote{Efficiency analysis of \OURS{} is in Appendix~\ref{app:efficiency_analysis}.}. The prompt used for extracting knowledge triples is provided in Appendix~\ref{app:prompt_knowledge_decomposition}, where the LLM is instructed to extract all knowledge triples contained within a document in a single pass. We denote the set of knowledge triples extracted from all the retrieved documents at step $i$ as $\tilde{\mathcal{T}^i}$. 

\vspace{0.5em} \noindent \textbf{Candidate Knowledge Identification}. 
To adapt the retrieval process to evolving information needs, \OURS{} retrieves a subset of candidate knowledge triples, i.e., $\mathcal{T}_q^i$, from all the extracted triples that are most likely to address the information gaps in the \jy{$i$-th step} reasoning chain. These candidate triples are selected based on their relevance to the question and their potential to form a coherent reasoning process with the \jy{$i$-th step} reasoning chain. 

To achieve this, we propose a \textit{Reasoning Chain Aligner}, which is designed to identify candidate triples that advance the reasoning process \jy{(see Step 1.2 in Figure~\ref{figure:model})}. 
We instantiate the Aligner as a bi-encoder model. At the $i$-th iteration, the Aligner encodes the query $q_i$, comprising the question and the $i$-th step reasoning chain, and each triple $t$ in $\tilde{\mathcal{T}^i}$ independently {into a shared space}. The score of each triple for addressing the information gaps in the $i$-th step reasoning chain is computed by taking the inner-product of the query and triple embeddings: 
$
    s_\theta(q_i, t) = f_\theta(q_i)^\top f_\theta(t), \ \forall t \in \tilde{\mathcal{T}^i}, 
$
where $f_\theta(\cdot)$ denotes the embedding function parameterised by $\theta$. 
The top-$N$\footnote{We provide analysis of the effect of $N$ in Appendix~\ref{app:effect_of_n}.} triples with the highest scores are selected as candidate triples to extend the $i$-th step reasoning chain, i.e., $\mathcal{T}_q^i$. The reasoning chain Aligner is trained to retrieve triples that contribute to building a coherent reasoning chain. Details of the training process are provided in \S\ref{subsec:training}. 

\vspace{0.5em} \noindent \textbf{Reasoning Chain Construction}. 
Given the candidate triples $\mathcal{T}_q^i$ from the Aligner at the $i$-th iteration, \OURS{} employs an LLM-based \textit{Reasoning Chain Constructor} to select a single triple from the candidates to extend the $i$-th step reasoning chain (see Step 1.3 in Figure~\ref{figure:model}). 
Our approach is inspired by IRCoT~\cite{trivedi2023interleaving}, which iteratively generates individual sentences in a chain-of-thought (CoT). However, instead of relying on potentially inaccurate CoTs, we instruct the LLM to generate a \text{chain-of-knowledge} (CoK)~\cite{wang2024boosting}, where free-form thoughts are replaced with document-grounded knowledge triples to ensure factual reliability.

The prompt used by the Constructor is provided in Appendix~\ref{app:prompt_reasoning_chain_construction}. The inputs include the question, the $i$-step reasoning chain and candidate triples $\mathcal{T}_q^i$.
The Constructor selects triples from \jy{$\mathcal{T}_q^i$ to complete the $i$-th step reasoning chain.} The first triple in the generated result is appended to the $i$-th step reasoning chain, forming a new chain that serves as input for subsequent iterations. Note that the Constructor aims to complete the whole chain, but we only take the first triple. Asking the Constructor to complete the whole chain reduces hallucination, and avoids a sub-optimal greedy approach.

The \jy{iterative process} terminates when the Constructor generates a reasoning chain containing ``\textit{the answer is}'' or reaches the maximum number of iterative steps $L$. The candidate knowledge triples collected during the iterative process, i.e., $\mathcal{T}_q = \{\mathcal{T}_q^i\}_{i=1}^L$, along with their associated scores, are output for document retrieval and ranking. 

\subsection{Document Ranking}
\label{subsec:document_ranking}
Since the \jy{retrieved} knowledge triples $\mathcal{T}_q$ may lack certain contextual information, we use these triples to identify and rank their source documents, i.e., $p(\mathcal{D}_q | q, \mathcal{T}_q)$ \jy{in Eq.~\ref{eq:kirag_retriever_model}}, to provide a more comprehensive and precise context. 
Specifically, the \jy{retrieved} documents $\mathcal{D}_q$ are collected by aggregating all the documents from which the triples in $\mathcal{T}_q$ are derived. To rank these documents, we assign each document the score of its associated triple(s) $s_\theta(q_i, t)$ from the iterative process. For a document associated with multiple triples, its score is determined by taking the highest one. These documents are ranked in descending order of their scores\jy{, with top-$K$ documents returned as the final retrieval results.}

Given the question $q$ and the ranked documents $\mathcal{D}_q$, \OURS{} leverages an LLM-based reader model to directly generate the answer. The prompt used for answer generation is provided in Appendix~\ref{app:prompt_answer_generation}, which instructs the model to leverage the context provided by the documents to answer the question. 

\subsection{Training Strategy}
\label{subsec:training}
\jy{In \OURS{}, the Reasoning Chain Aligner is the key component that requires training to effectively identify candidate triples for extending reasoning chain, while the other components, i.e., Retriever and Constructor, remain frozen.}
This section outlines the training strategy for the Aligner. 
Due to the lack of existing datasets specifically designed for this task, we construct a silver training dataset by adapting data from existing multi-hop QA datasets. 
\jy{Specifically, given a question and its ground-truth relevant documents, we construct a knowledge triple-based reasoning chain that supports answering the question. The reasoning chain and the question will serve as the labeled data for training the Aligner.}

To train the Aligner, we decompose the complete reasoning chain into multiple incomplete reasoning chains and the corresponding next triples (see \jy{the right part of} Figure~\ref{figure:model}). For each incomplete reasoning chain, the correct next triple is treated as the positive sample, while the other triples from the candidate set $\tilde{\mathcal{T}}^i$ are treated as negative samples. The aligner is trained with contrastive learning loss:
\vspace{-1.5em}
\begin{align}
    \mathcal{L} = - \hspace{-1.2em} \sum\limits_{(q, r, t^+) \in \mathcal{P}} \hspace{-1em} \log \frac{g_\theta(q_r, t^+)}{g_\theta(q_r, t^+) + \hspace{-0.5em}\sum\limits_{t^- \in \tilde{\mathcal{T}}^{|r|}} \hspace{-0.5em}g_\theta(q_r, t^-)},
\end{align}
where $\mathcal{P}$ is the training set. Each data-point includes a question $q$, an incomplete reasoning chain $r$ and a positive triple $t^+$. The query $q_r$ is the concatenation of $q$ and $r$, and the function $g_\theta(q_r, t)=\exp(s_\theta(q_r, t))/\tau$ computes the logits, with $\tau$ being the temperature. 
\jy{Further details on the training data and training process are provided in Appendix}~\ref{app:training_hparam_details}.

\section{Experiments}

\subsection{Experimental Setup}

\noindent \textbf{Datasets}. 
We conduct experiments on five multi-hop QA datasets: 
\textbf{HotPotQA}~\cite{yang2018hotpotqa}, \textbf{2WikiMultiHopQA} (\textbf{2Wiki})~\cite{ho2020constructing}, \textbf{MuSiQue}~\cite{trivedi2022musique}, \textbf{Bamboogle}~\cite{press2023measuring} and \textbf{WebQuestions} (\textbf{WebQA})~\cite{berant2013semantic}. 
We also use a single-hop QA dataset: \textbf{Natural Questions} (\textbf{NQ})~\cite{kwiatkowski2019natural}. 
We report the performance on the \textit{full} test sets of these datasets. For datasets with non-pulic test sets (HotPotQA, 2Wiki and MuSiQue), we use their development sets as test sets and report corresponding results. Detailed statistics and corpus information are provided in Appendix~\ref{app:datasets}. 

\vspace{0.5em} \noindent \textbf{Baselines}. 
Since \OURS{} aims to improve the retrieval performance of \jy{iRAG models,} we primarily compare it with \jy{iRAG} models. We compare \OURS{} with models from the following categories: (1) Standard RAG model; 
(2) \jy{iRAG models,} such as IRCoT~\cite{trivedi2023interleaving}, FLARE~\cite{jiang2023active}, and DRAGIN~\cite{su2024dragin}; 
(3) Enhanced retrieval models, which improve the retrieval performance by using feedback from earlier retrieval steps, such as BeamDR~\cite{zhao2021multi} and Vector-PRF~\cite{li2023pseudo}. 
Moreover, to evaluate the effectiveness of using knowledge triples for iterative retrieval, we introduce two variants: \textit{\OURS{}-Doc} and \textit{\OURS{}-Sent}, where the triples are replaced with documents and sentences, respectively. Both variants follow the same procedure as \OURS{} to retrieve documents. 
More details about the baselines can be found in Appendix~\ref{app:baselines}. 

\vspace{0.5em} \noindent \textbf{Evaluation}. 
To evaluate the retrieval performance, we follow previous works~\cite{trivedi2023interleaving,gutierrez2024hipporag} and use \textbf{R$\text{@}$\{3, 5\}} as the metrics. To evaluate the QA performance, we use \textbf{Exact Match} (\textbf{EM}) and \textbf{F1} as evaluation metrics, which are the standard metrics for these datasets.

\begin{table}[tb]
\centering
\resizebox{0.48\textwidth}{!}{
\begin{tabular}{lcccccc}
\toprule
\multirow{2}{*}{\textbf{Model}} & \multicolumn{2}{c}{\textbf{HotPotQA}} & \multicolumn{2}{c}{\textbf{2Wiki}} & \multicolumn{2}{c}{\textbf{MuSiQue}}\\
\cmidrule(lr){2-3} \cmidrule(lr){4-5} \cmidrule(lr){6-7}
& \textbf{R}$\text{@}$\textbf{3} & \textbf{R}$\text{@}$\textbf{5} & \textbf{R}$\text{@}$\textbf{3} & \textbf{R}$\text{@}$\textbf{5} & \textbf{R}$\text{@}$\textbf{3} & \textbf{R}$\text{@}$\textbf{5} \\
\midrule
\textbf{RAG} & 65.47\textcolor{white}{$^*$} & 70.78\textcolor{white}{$^*$} & 60.87\textcolor{white}{$^*$} & 65.20\textcolor{white}{$^*$} & 41.29\textcolor{white}{$^*$} & 46.53\textcolor{white}{$^*$} \\
\hdashline\noalign{\vskip 0.5ex}
\textbf{Vector-PRF} & 65.37\textcolor{white}{$^*$} & 70.06\textcolor{white}{$^*$} & 60.60\textcolor{white}{$^*$} & 64.85\textcolor{white}{$^*$} & 40.93\textcolor{white}{$^*$} & 45.46\textcolor{white}{$^*$}  \\
\textbf{BeamDR} & 67.07\textcolor{white}{$^*$} & 71.89\textcolor{white}{$^*$} & 36.07\textcolor{white}{$^*$} & 42.08\textcolor{white}{$^*$} & 24.17\textcolor{white}{$^*$} & 28.18\textcolor{white}{$^*$} \\
\hdashline\noalign{\vskip 0.5ex}
\textbf{FLARE} & 54.79\textcolor{white}{$^*$} & 59.72\textcolor{white}{$^*$} & 60.84\textcolor{white}{$^*$} & 70.04\textcolor{white}{$^*$} & 39.79\textcolor{white}{$^*$} & 45.81\textcolor{white}{$^*$} \\
\textbf{DRAGIN} & 69.95\textcolor{white}{$^*$} & 75.85\textcolor{white}{$^*$} & 61.30\textcolor{white}{$^*$} & 70.43\textcolor{white}{$^*$} & \underline{48.67}\textcolor{white}{$^*$} & \underline{54.67}\textcolor{white}{$^*$} \\
\textbf{IRCoT} & \underline{71.44}\textcolor{white}{$^*$} & \underline{77.57} \textcolor{white}{$^*$} & \underline{64.30}\textcolor{white}{$^*$} & \underline{75.56}\textcolor{white}{$^*$}  &45.61\textcolor{white}{$^*$} &52.21\textcolor{white}{$^*$} \\
\hdashline\noalign{\vskip 0.5ex}
\textbf{\OURS{}-Doc}
& 67.80\textcolor{white}{$^*$} & 72.20\textcolor{white}{$^*$} & 45.85\textcolor{white}{$^*$} & 63.07\textcolor{white}{$^*$} & 25.86\textcolor{white}{$^*$} & 39.49\textcolor{white}{$^*$}  \\
\textbf{\OURS{}-Sent}
& 54.43\textcolor{white}{$^*$} & 69.26\textcolor{white}{$^*$} & 43.53\textcolor{white}{$^*$} & 59.33\textcolor{white}{$^*$} & 31.08\textcolor{white}{$^*$} & 43.69\textcolor{white}{$^*$} \\
\textbf{\OURS{}} &\textbf{80.32}$^\dagger$ & \textbf{84.08}$^\dagger$ & \textbf{77.76}$^\dagger$ & \textbf{85.32}$^\dagger$ & \textbf{54.53}$^\dagger$ & \textbf{61.16}$^\dagger$ \\

\bottomrule    
\end{tabular}
}
\vspace{-0.5em}
\caption{Retrieval performance (\%) on multi-hop QA datasets, with the best and second-best results marked in bold and underlined, respectively, and $^\dagger$ denotes p-value<0.05 compared with best-performing baseline. }
\vspace{-0.5em}
\label{table:in_domain_retrieval_performance}
\end{table}

\begin{table}[tb]
\centering
\resizebox{0.48\textwidth}{!}{
\begin{tabular}{lcccccc}
\toprule
\multirow{2}{*}{\textbf{Model}} & \multicolumn{2}{c}{\textbf{HotPotQA}} & \multicolumn{2}{c}{\textbf{2Wiki}} & \multicolumn{2}{c}{\textbf{MuSiQue}}\\
\cmidrule(lr){2-3} \cmidrule(lr){4-5} \cmidrule(lr){6-7}
& \textbf{EM} & \textbf{F1} & \textbf{EM} & \textbf{F1} & \textbf{EM} & \textbf{F1} \\
\midrule
\textbf{RAG} & 34.54\textcolor{white}{$^*$} & 47.35\textcolor{white}{$^*$} & 14.78\textcolor{white}{$^*$} & 30.48\textcolor{white}{$^*$} & \textcolor{white}{0}9.10\textcolor{white}{$^*$} & 16.98\textcolor{white}{$^*$} \\
\hdashline\noalign{\vskip 0.5ex}
\textbf{Vector-PRF} 
& 34.40\textcolor{white}{$^*$} & 47.31\textcolor{white}{$^*$} & 14.96\textcolor{white}{$^*$} & 30.37\textcolor{white}{$^*$} & \textcolor{white}{0}9.23\textcolor{white}{$^*$} & 16.98\textcolor{white}{$^*$} \\
\textbf{BeamDR} & 38.34\textcolor{white}{$^*$} & 51.64\textcolor{white}{$^*$} & 14.42\textcolor{white}{$^*$} & 27.25\textcolor{white}{$^*$} & \textcolor{white}{0}7.08\textcolor{white}{$^*$} & 14.42\textcolor{white}{$^*$}  \\ 
\hdashline\noalign{\vskip 0.5ex}
\textbf{FLARE} & 35.58\textcolor{white}{$^*$} & 47.74\textcolor{white}{$^*$} & \underline{26.36}\textcolor{white}{$^*$} & \underline{41.82}\textcolor{white}{$^*$} & 13.07\textcolor{white}{$^*$} & 21.94\textcolor{white}{$^*$} \\
\textbf{DRAGIN} & 41.74\textcolor{white}{$^*$} & 55.69\textcolor{white}{$^*$} & {25.58}\textcolor{white}{$^*$} & 40.83\textcolor{white}{$^*$} & \underline{16.87}\textcolor{white}{$^*$} & \underline{26.71}\textcolor{white}{$^*$} \\
\textbf{IRCoT} & \underline{42.38}\textcolor{white}{$^*$} & \underline{56.38}\textcolor{white}{$^*$} & 25.12\textcolor{white}{$^*$} & {41.36}\textcolor{white}{$^*$} & 15.76\textcolor{white}{$^*$} & 24.94\textcolor{white}{$^*$} \\
\hdashline\noalign{\vskip 0.5ex}
\textbf{\OURS{}-Doc} 
& 33.87\textcolor{white}{$^*$} & 46.43\textcolor{white}{$^*$} & 14.37\textcolor{white}{$^*$} & 27.54\textcolor{white}{$^*$} & \textcolor{white}{0}7.49\textcolor{white}{$^*$} & 15.34\textcolor{white}{$^*$} \\
\textbf{\OURS{}-Sent}  
& 34.14\textcolor{white}{$^*$} & 46.63\textcolor{white}{$^*$} & 14.22\textcolor{white}{$^*$} & 27.50\textcolor{white}{$^*$} & 10.51\textcolor{white}{$^*$} & 18.21\textcolor{white}{$^*$}  \\
\textbf{\OURS{}} 
&\textbf{45.09}\textcolor{black}{$^\dagger$} & \textbf{59.76}$^\dagger$ & \textbf{30.72}$^\dagger$ & \textbf{50.57}$^\dagger$ & \textbf{19.16}{$^\dagger$} & \textbf{30.00}$^\dagger$ \\
\bottomrule    
\end{tabular}
}
\vspace{-0.5em}
\caption{QA performance (\%) on multi-hop QA datasets, with the best and second-best results marked in bold and underlined, respectively. \hspace{-0.2em}$^\dagger$ denotes p-value<0.05 compared with best-performing baseline. }
\vspace{-0.7em}
\label{table:in_domain_qa_performance}
\end{table}

\vspace{0.5em} \noindent \textbf{Training and Implementation Details}. 
To train the Aligner, we use TRACE~\cite{fang2024trace}, which constructs knowledge triple-based reasoning chains from a fixed set of documents, to generate ground-truth reasoning chains. The reasoning chain that leads to the correct answer is used for training. Training data is generated from the training sets of three multi-hop QA datasets: HotPotQA, 2Wiki and MuSiQue. The combined data is used to train the Aligner. The Aligner is initialised with E5~\cite{wang2022text} and finetuned with the \jy{constructed} training data. 

\OURS{} uses Llama3~\cite{dubey2024llama} to extract triples and serve as the Constructor to build reasoning chains. It uses frozen E5 or BGE~\cite{xiao2023c} \jy{as the Retriever.} We use different readers, including Llama3, Qwen2.5~\cite{yang2024qwen2}, Flan-T5~\cite{chung2024scaling} and TRACE~\cite{fang2024trace} to generate answers. We mainly report results using E5 as the retriever and Llama3 as the reader, with additional results from other retrievers and readers provided in Appendix~\ref{app:overall_performance_different_retrievers_readers}. For fair comparison, RAG baselines employ the same retriever and reader as \OURS{}. More training and implementation details are in Appendix~\ref{app:training_hparam_details}.

\begin{table}[tb]
\centering
\vspace{-0.8em}
\resizebox{0.48\textwidth}{!}{
\begin{tabular}{lcccccc}
\toprule
\multirow{2}{*}{\textbf{Model}} & \multicolumn{2}{c}{\textbf{Bamboogle}} & \multicolumn{2}{c}{\textbf{WebQA}} & \multicolumn{2}{c}{\textbf{NQ}}\\
\cmidrule(lr){2-3} \cmidrule(lr){4-5} \cmidrule(lr){6-7}
& \textbf{R}$\text{@}$\textbf{3} & \textbf{R}$\text{@}$\textbf{5} & \textbf{R}$\text{@}$\textbf{3} & \textbf{R}$\text{@}$\textbf{5} & \textbf{R}$\text{@}$\textbf{3} & \textbf{R}$\text{@}$\textbf{5} \\
\midrule
\textbf{RAG} & 20.80\textcolor{white}{$^*$} & 25.60\textcolor{white}{$^*$} & 64.91\textcolor{white}{$^*$} & 70.32\textcolor{white}{$^*$} & \underline{73.07}\textcolor{white}{$^*$} & \underline{78.56}\textcolor{white}{$^*$}  \\
\hdashline\noalign{\vskip 0.5ex}

\textbf{Vector-PRF} 
& 20.60\textcolor{white}{$^*$} & 24.80\textcolor{white}{$^*$} & 64.86\textcolor{white}{$^*$} & 69.54\textcolor{white}{$^*$} & 72.82\textcolor{white}{$^*$} & 78.03\textcolor{white}{$^*$}  \\

\textbf{BeamDR} & 12.00\textcolor{white}{$^*$} & 15.20\textcolor{white}{$^*$} & 41.63\textcolor{white}{$^*$} & 50.25\textcolor{white}{$^*$} & 33.88\textcolor{white}{$^*$} & 42.16\textcolor{white}{$^*$}   \\ 
\hdashline\noalign{\vskip 0.5ex}

\textbf{FLARE} & 32.80\textcolor{white}{$^*$} & 37.60\textcolor{white}{$^*$} & 55.91\textcolor{white}{$^*$} & 60.97\textcolor{white}{$^*$} & 68.98\textcolor{white}{$^*$} & 73.43\textcolor{white}{$^*$}  \\
\textbf{DRAGIN} & \underline{36.80}\textcolor{white}{$^*$} & \underline{40.40}\textcolor{white}{$^*$} & 65.11\textcolor{white}{$^*$} & 70.03\textcolor{white}{$^*$} & 68.98\textcolor{white}{$^*$} & 73.43\textcolor{white}{$^*$} \\
\textbf{IRCoT} & 28.00\textcolor{white}{$^*$} & 32.80\textcolor{white}{$^*$} & \underline{65.50}\textcolor{white}{$^*$} & \underline{70.42}\textcolor{white}{$^*$} & \textbf{73.38}\textcolor{white}{$^*$} & \textbf{78.59}\textcolor{white}{$^*$} \\
\hdashline\noalign{\vskip 0.5ex}

\textbf{\OURS{}-Doc}
& 20.80\textcolor{white}{$^*$} & 27.20\textcolor{white}{$^*$} & 62.40\textcolor{white}{$^*$} & 68.60\textcolor{white}{$^*$} & 68.59\textcolor{white}{$^*$} & 74.99\textcolor{white}{$^*$} \\ 

\textbf{\OURS{}-Sent}
& 26.40\textcolor{white}{$^*$} & 32.00\textcolor{white}{$^*$} & 62.16\textcolor{white}{$^*$} & 68.06\textcolor{white}{$^*$} & 67.48\textcolor{white}{$^*$} & 74.13\textcolor{white}{$^*$}  \\

\textbf{\OURS{}}  & \textbf{45.60}$^\dagger$ & \textbf{49.60}$^\dagger$ & \textbf{69.05}$^\dagger$ & \textbf{73.08}$^\dagger$ & 72.11\textcolor{white}{$^*$} & 77.28\textcolor{white}{$^\dagger$} \\
\bottomrule    
\end{tabular}
}
\vspace{-0.5em}
\caption{Retrieval performance (\%) on unseen multi-hop and single-hop QA datasets, where $^\dagger$ denotes p-value<0.05 compared with best-performing baselines.}
\vspace{-0.5em}
\label{table:unseen_retrieval_performance}
\end{table}

\subsection{Results and Analysis}
We present our primary results in this section. Additional results are provided in Appendix~\ref{app:additional_results_and_analysis}. 

\vspace{0.5em} \noindent \textbf{(RQ1): How does \OURS{} perform in multi-hop QA compared with baselines?}
The retrieval and QA\footnote{QA performance is based on the top-3 retrieved documents. The results for the top-5 retrieved documents are provided in Appendix~\ref{app:qa_performance_top_5_docs}, which demonstrate similar results.} results are provided in Table~\ref{table:in_domain_retrieval_performance} and Table~\ref{table:in_domain_qa_performance}, respectively, which yield the following findings: 

\noindent (1) \OURS{} consistently outperforms all baselines in retrieval performance on all datasets. Compared to the strongest baselines, \OURS{} achieves statistically significant average improvements of 9.40\% in R$\text{@}$3 and 7.59\% in R$\text{@}$5, demonstrating its superior ability to enhance retrieval performance.  

\noindent (2) \OURS{} consistently achieves the best QA performance on all datasets. It significantly outperforms best-performing baselines, with average improvements of 3.12\% in EM and 5.14\% in F1. 
The results validate the effectiveness of \OURS{} in facilitating multi-hop QA through high-quality retrieval. 

\noindent (3) Compared to \OURS{}-Doc and \OURS{}-Sent, which perform iterative retrieval at document and sentence levels, \OURS{} achieves substantially higher retrieval performance. 
The suboptimal performance of these variants stems from the iterative retrieval process being misled by noise in documents and sentences. In contrast, \OURS{} uses finer-grained knowledge triples, reducing the impact of noise and improving retrieval recall. 

\noindent (4) Compared to IRCoT, which uses CoT for iterative retrieval, \OURS{} achieves superior retrieval results, with an average improvement of 10.42\% in R$\text{@}$3. This improvement stems from LLM's tendency to generate hallucinated CoT. By using document-grounded knowledge triples, \OURS{} ensures a more reliable and faithful retrieval process. 

\begin{table*}[tb]
\begin{center}
\vspace{-1.0em}
\resizebox{0.99\textwidth}{!}{
\begin{tabular}{lccccccccccccc}
\toprule
\multirow{2}{*}{\textbf{Model}}
& \multirow{2}{*}{\textbf{Retriever}} 
& \multirow{2}{*}{\textbf{Aligner}} 
& \multirow{2}{*}{\textbf{Constructor}}
& \multicolumn{3}{c}{\textbf{HotPotQA}} 
& \multicolumn{3}{c}{\textbf{2Wiki}} 
& \multicolumn{3}{c}{\textbf{MuSiQue}} 
\\
\cmidrule(lr){5-7} \cmidrule(lr){8-10} \cmidrule(lr){11-13}
& & & &\textbf{R}$\text{@}$\textbf{3} & \textbf{R}$\text{@}$\textbf{5} & \textbf{F1} & \textbf{R}$\text{@}$\textbf{3} & \textbf{R}$\text{@}$\textbf{5} &\textbf{F1} & \textbf{R}$\text{@}$\textbf{3} & \textbf{R}$\text{@}$\textbf{5} & \textbf{F1} \\
\midrule

\textbf{\OURS{}} & {\Large \ding{51}} & {\Large \ding{51}}({trained}) & {\Large \ding{51}} 
& \textbf{80.32}\textcolor{white}{$^*$} & \textbf{84.08}\textcolor{white}{$^*$} & \textbf{59.76}\textcolor{white}{$^*$} & \textbf{77.76}\textcolor{white}{$^*$} & \textbf{85.32}\textcolor{white}{$^*$} & \textbf{50.57}\textcolor{white}{$^*$} & \textbf{54.53}\textcolor{white}{$^*$} & \textbf{61.16}\textcolor{white}{$^*$} & \textbf{30.00}\textcolor{white}{$^*$} \\

\noalign{\vskip 0.5ex}\hdashline\noalign{\vskip 0.5ex} 
\textbf{\OURS{} w/o Retriever} 
& {\Large \ding{55}} & {\Large \ding{51}}({trained}) & {\Large \ding{51}} 
& 74.09$^\dagger$  & 78.03$^\dagger$ & 57.07$^\dagger$  & 77.29\textcolor{white}{$^\dagger$} & 83.65$^\dagger$ & 48.76$\dagger$  & 53.93\textcolor{white}{$^\dagger$}  & 60.68\textcolor{white}{$^\dagger$} & 27.14$^\dagger$ \\

\textbf{\OURS{} w/o Aligner} 
& {\Large \ding{51}}  & {\Large \ding{55}} & {\Large \ding{51}}
& 73.34$^\dagger$ & 75.79$^\dagger$ & 53.47$^\dagger$ & 67.66$^\dagger$ & 70.73$^\dagger$ & 39.06$^\dagger$ & 45.60$^\dagger$ & 49.62$^\dagger$ & 21.80$^\dagger$ \\

\textbf{\OURS{} w/o Constructor} 
& {\Large \ding{51}} & {\Large \ding{51}}({trained}) & {\Large \ding{55}} 
& 74.96$^\dagger$ & 79.51$^\dagger$	& 55.67$^\dagger$ & 72.64$^\dagger$ & 80.89$^\dagger$	& 45.69$^\dagger$ & 46.98$^\dagger$	& 55.12$^\dagger$ & 23.71$^\dagger$ \\

\textbf{\OURS{} w/o Training} & {\Large \ding{51}} & {\Large \ding{51}}({w/o training}) & {\Large \ding{51}}
& 76.35{$^\dagger$} & 81.56{$^\dagger$} & {59.03}{$^\dagger$} & {75.37}{$^\dagger$} & {82.50}{$^\dagger$} & {48.33}{$^\dagger$} & {51.33}{$^\dagger$} & {58.66}{$^\dagger$} & {28.08}{$^\dagger$} \\

\bottomrule
\end{tabular}
}
\end{center}
\vspace{-1.0em}
\caption{Ablation studies of \OURS{}, where $^\dagger$ indicates p-value < 0.05 compared with \OURS{}.}
\vspace{-0.8em}
\label{table:ablation_study}
\end{table*}

\begin{figure}[tb]
\begin{center}
\vspace{-0.8em}
\includegraphics[width=0.48\textwidth]{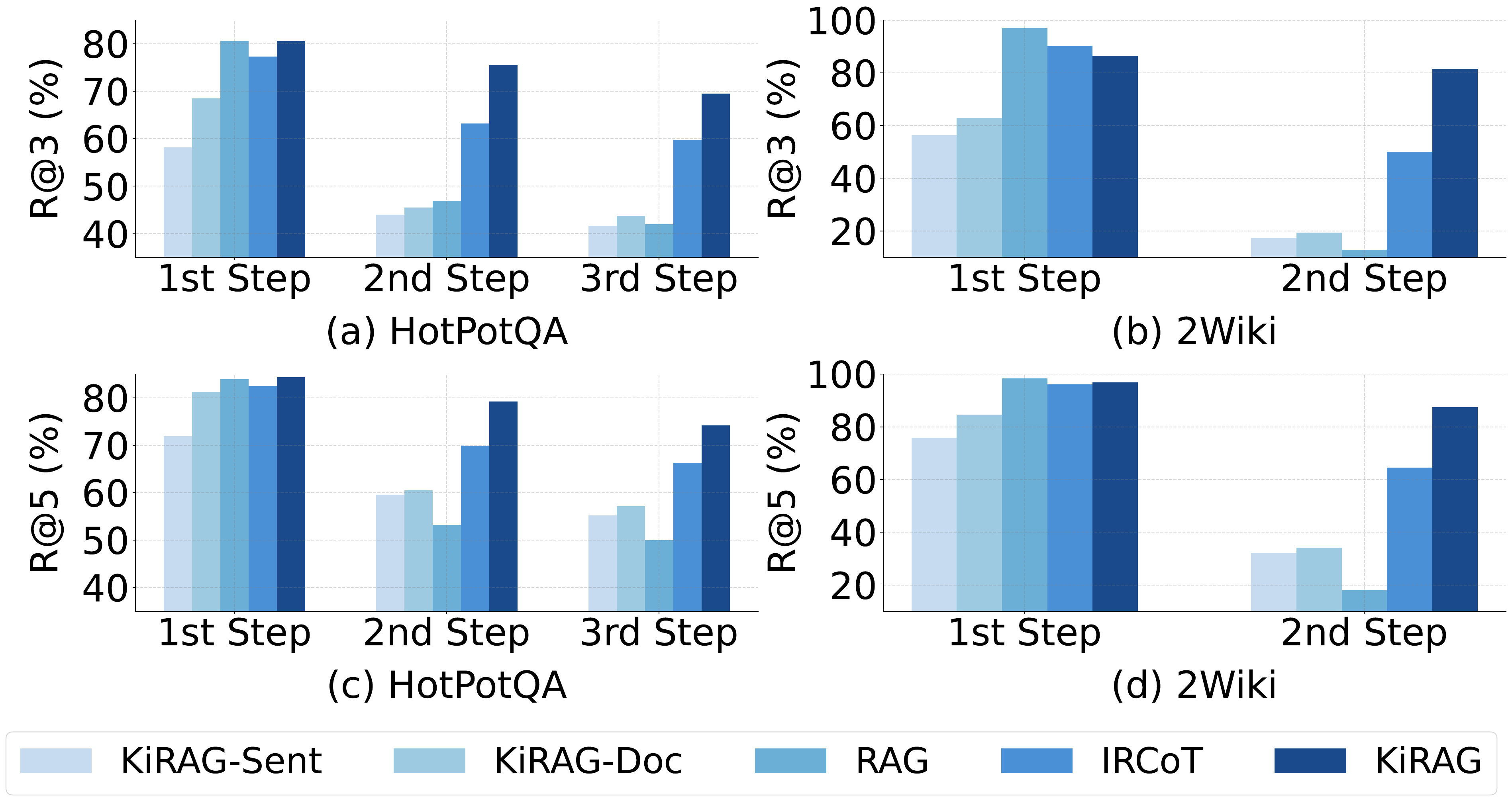}
\end{center}
\vspace{-0.8em}
\caption{Retrieval performance (\%) \jy{for relevant documents} required across different steps, where most questions in 2Wiki have only two relevant documents.}
\label{figure:op_recall_different_hop_hotpot_wiki}
\vspace{-1.0em}
\end{figure}

\vspace{0.5em} \noindent \textbf{(RQ2): Why does \OURS{} improve retrieval performance for multi-hop questions?}
To explain why \OURS{} achieves superior retrieval performance, we analyze its ability to retrieve relevant documents required at different steps of the reasoning process. 
For a multi-hop question, there are multiple logically ordered relevant documents. 
For instance, the first relevant document for the question in Figure~\ref{figure:model} is about ``Kirton End'', while the second relevant document relates to ``Boston''.  
At step $i$, \jy{we only consider the document required at that specific step as relevant and compute its recall}. This approach allows us to assess how well \OURS{} retrieves the necessary information at each step. 

The results on HotPotQA and 2Wiki are shown in Figure~\ref{figure:op_recall_different_hop_hotpot_wiki}, yielding the following findings: 
(1) The recall of both \OURS{} and baselines declines with increasing steps, highlighting the growing challenge of retrieving relevant documents for later steps in the reasoning process;  
(2) \jy{Compared to RAG and IRCoT}, \OURS{} shows comparable, and occasionally slightly lower, retrieval recall at the first step. However, it achieves substantially higher retrieval recall in subsequent steps, \jy{which contributes to its overall retrieval effectiveness. This improvement stems from \OURS{}'s iterative retrieval process, which dynamically adapts to evolving information needs, enabling the effective retrieval of relevant documents required at each step.}

\vspace{0.5em} \noindent \textbf{(RQ3): Can \OURS{} effectively generalise to unseen multi-hop and single-hop QA datasets?}
To evaluate the generalisation ability of \OURS{}, we conduct additional experiments on two multi-hop QA datasets, Bamboogle and WebQA, as well as a single-hop QA dataset, NQ, none of which were included during training. The retrieval results\footnote{The QA performance, presented in Table~\ref{table:unseen_qa_performance} of the Appendix, shows consistent results with retrieval performance.} are presented in Table~\ref{table:unseen_retrieval_performance}, which shows that \OURS{} significantly outperforms all baselines on two multi-hop QA datasets, and demonstrates comparable retrieval performance to IRCoT, the best-performing baseline, on the single-hop QA dataset NQ. These findings highlight the strong generalisation ability of \OURS{} in handling diverse QA tasks.

\begin{table}[tb]
\centering
\resizebox{0.48\textwidth}{!}{
\begin{tabular}{p{4.5cm}cccc}
\toprule
\multirow{2}{*}{\textbf{Model}} & \multicolumn{2}{c}{\textbf{Retrieval}} & \multicolumn{2}{c}{\textbf{QA}} \\
\cmidrule(lr){2-3}\cmidrule(lr){4-5}
& \textbf{R$\text{@}$3} & \textbf{R$\text{@}$5} & \textbf{EM} & \textbf{F1} \\
\midrule
\textbf{E5} & 29.50\textcolor{white}{$^*$} & 43.25\textcolor{white}{$^*$} & 23.00\textcolor{white}{$^*$} & 31.56\textcolor{white}{$^*$}  \\
\textbf{\OURS{} w/o Constructor} & 76.50$^\dagger$ & 79.25$^\dagger$ & 31.00$^\dagger$ & 48.22$^\dagger$ \\ 
\textbf{\OURS{}} 
& \textbf{84.25}$^\dagger$ & \textbf{86.50}$^\dagger$ & \textbf{39.00}$^\dagger$ & \textbf{53.63}$^\dagger$ \\
\bottomrule
\end{tabular}
}
\vspace{-0.5em}
\caption{Performance in retrieving relevant knowledge triples for the 100 manually labeled questions on 2Wiki, where $^\dagger$ denotes p-value$<$0.05 compared with E5.}
\vspace{-1.0em}
\label{table:triple_retrieval_effectiveness}
\end{table}

\vspace{0.5em} \noindent \textbf{(RQ4): What are the effects of each component and the training strategy in \OURS{}?} 
To evaluate the impact of the Retriever, we introduce \textit{\OURS{} w/o Retriever}, where the retriever is removed and candidate triples are directly retrieved from the knowledge graph corpus using the Reasoning Chain Aligner. Table~\ref{table:ablation_study} shows that removing the Retriever leads to a significant performance drop on HotPotQA while maintaining comparable performance on 2Wiki and MuSiQue. 
This demonstrates the Aligner's effectiveness in identifying relevant knowledge triples but highlights the limitations of relying solely on the knowledge graph, which may loss contextual information present in documents. 

To assess the impact of the Aligner, we introduce \textit{\OURS{} w/o Aligner}, where the Aligner is removed and all knowledge triples from the retrieved documents are passed to the Reasoning Chain Constructor. 
Table~\ref{table:ablation_study} shows that \textit{\OURS{} w/o Aligner} suffers an average decrease of 8.67\% in R$\text{@}$3 and 11.47\% in R$\text{@}$5 compared to \OURS{}. 
This decline is due to the absence of filtering or ranking by the Aligner, resulting in noisy and irrelevant triples that hinder the Reasoning Chain Constructor's ability to build coherent reasoning chains, which is essential for guiding the iterative retrieval process effectively. 

Moreover, to evaluate the impact of the Constructor, we introduce \textit{\OURS{} w/o Constructor}, which constructs reasoning chain using only the top-ranked triple identified by the Aligner. Table~\ref{table:ablation_study} indicates that removing the Constructor leads to significantly inferior performance, highlighting the importance of the LLM-based Constructor in building coherent reasoning chains through its advanced reasoning and contextual understanding capability. 

\begin{figure}[tb]
\begin{center}
\includegraphics[width=0.45\textwidth]{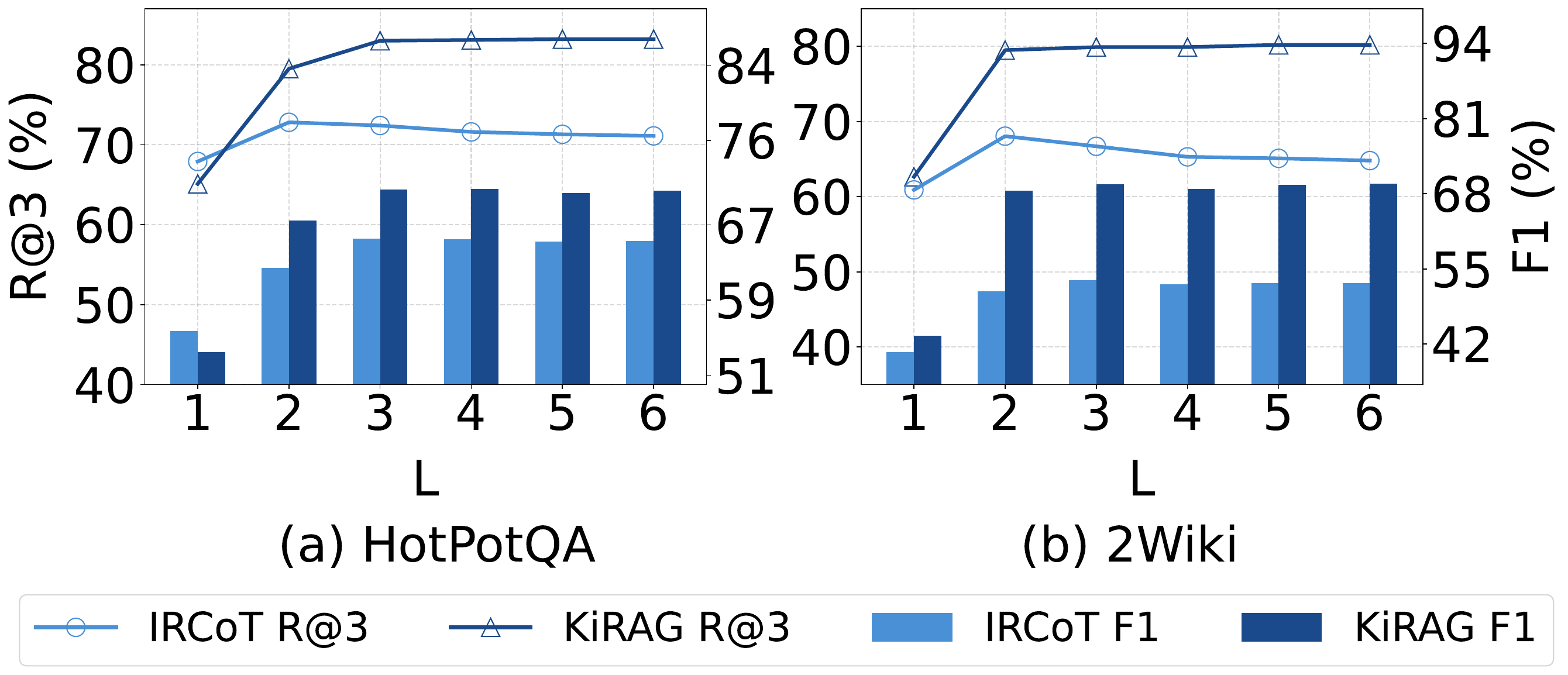}
\end{center}
\vspace{-0.8em}
\caption{The effect of the number of iterative steps $L$.}
\label{figure:effect_num_turns}
\vspace{-1.0em}
\end{figure}

\jy{To assess the impact of training the Aligner for retrieving and integrating triples, we introduce \textit{\OURS{} w/o Training}, where the Aligner is replaced with a frozen E5, which is trained for general text retrieval. Table~\ref{table:ablation_study} shows that \textit{\OURS{} w/o Training} exhibits a significant decline in both retrieval and QA results.} 
These results highlight the effectiveness of our training strategy in enabling the Aligner to identify relevant knowledge triples. 

\vspace{0.5em} \noindent \textbf{(RQ5): Can \OURS{} retrieve relevant knowledge triples to address multi-hop questions?} 
To evaluate the quality of knowledge triples retrieved by \OURS{}, we randomly select 100 questions from the 2Wiki test set and manually identify knowledge triples that are useful in answering these questions. These manually selected triples are considered relevant\footnote{\jy{Appendix~\ref{app:details_examples_relevant_triples} provides details and examples of the manually curated data, which will be released alongside the code.}}. 
We use R$\text{@}$K to measure retrieval performance and compute QA metrics (EM and F1) using the retrieved triples as context. \jy{We compare \OURS{} with E5}, which directly retrieves knowledge triples from the knowledge graph corpus, and \jy{the Reasoning Chain Aligner, which iteratively retrieves triples using the trained Aligner.} Table~\ref{table:triple_retrieval_effectiveness} shows that \OURS{} significantly outperforms E5 in both retrieval and QA performance, demonstrating its effectiveness in retrieving relevant knowledge triples. This superior performance is attributed to the Aligner and the Constructor, which are specifically designed to identify relevant triples.

\vspace{0.5em} \noindent \textbf{(RQ6): How does the number of iterative steps $L$ affect the performance of \OURS{}?} 
Table~\ref{figure:effect_num_turns} shows the retrieval and QA results of \OURS{} with different values of $L$ on HotPotQA and 2Wiki development sets. 
The results show that as the value of $L$ increases, both retrieval and QA performance initially improve and then reach a plateau, with \OURS{} achieving optimal performance at a moderate value of $L$. This highlights the importance of selecting a proper value of $L$ in \OURS{} to balance high accuracy and efficiency in multi-hop QA.

\begin{figure}[tb]
\begin{center}
\includegraphics[width=0.45\textwidth]{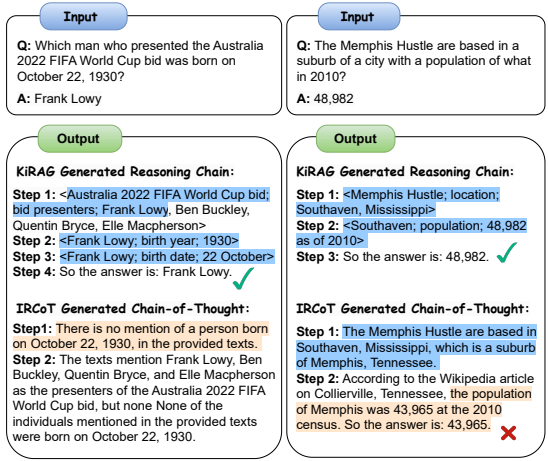}
\end{center}
\vspace{-0.8em}
\caption{Case study of \OURS{} and IRCoT on HotPotQA test set, where the relevant and irrelevant context are marked in \colorbox{myblue}{blue} and \colorbox{myorange}{orange}, respectively.}
\label{figure:case_study}
\vspace{-1.0em}
\end{figure}

\vspace{0.5em} \noindent \textbf{Case Study.} 
We conduct a case study to examine the reasoning chains generated by \OURS{}. 
Figure~\ref{figure:case_study} shows examples of the reasoning chains produced by \OURS{} and the CoTs generated by IRCoT. 
The examples show that \OURS{} can generate coherent and contextually relevant reasoning chains for answering multi-hop questions, which are essential for effectively guiding the iterative retrieval process. 
In contrast, IRCoT may struggle with missing information or hallucinations, hindering its ability to retrieve the necessary knowledge.

\section{Related Work}

\textbf{RAG Models.} 
RAG models have shown superior performance in QA tasks~\cite{lewis2020retrieval,izacard2020leveraging,ram2023context}. 
These models typically employ the \textit{retriever-reader} architecture, which consists of a retriever~\cite{karpukhin2020dense,wang2022text,fang2023kgpr} % for retrieving relevant documents 
and a reader~\cite{izacard2020leveraging,jiang2023active}. 
Efforts to improves RAG models generally follows three main directions: 
(1) enhance the retriever for better retrieval performance~\cite{izacard2021distilling,shi2023replug,wang2024richrag}; 
(2) enhance the reader for better comprehension and answer generation~\cite{lin2024ra,xu2024unsupervised,wang2024rear}; 
(3) introduce additional modules to bridge the retriever and the reader~\cite{yu2023generate,xu2023recomp,ye2024r}. 

\vspace{0.5em} \noindent \textbf{Iterative RAG Models for Multi-Hop QA.}
Iterative RAG models~\cite{trivedi2023interleaving,shao2023enhancing,asai2024self,liu2024ra,yao2024seakr} address multi-hop QA by performing multiple steps of retrieval and reasoning. 
For instance, IRCoT~\cite{trivedi2023interleaving} use LLM-generated chain-of-thoughts for retrieval, while  DRAGIN~\cite{su2024dragin} dynamically decides when and what to retrieve based on the LLM's information needs. 
However, these models all rely on LLM-generated thoughts, making them prone to hallucination. In contrast, \OURS{} employs knowledge triples and a trained retriever to actively identify and retrieve missing information, enabling a more reliable and accurate retrieval for multi-hop QA. 

\vspace{0.5em} \noindent \textbf{KG-Enhanced RAG Models.} 
Recently, KGs have been integrated into RAG models~\cite{peng2024graph}. Some studies leverage information from existing KGs~\cite{vrandevcic2014wikidata} for additional context~\cite{yu2022kg,sun2024think}, while others generate KGs from documents to improve knowledge organisation~\cite{edge2024local,gutierrez2024hipporag,chen2024kg} or enhance reader comprehension~\cite{li2023leveraging,fang2024reano,fang2024trace,panda2024holmes}. 
These models primarily follow the standard RAG pipeline, whereas our work focuses on the iRAG pipelines. Moreover, while they rely on single-step retrieval with pre-existing retrievers, \OURS{} employs a trained retriever tailored for iterative retrieval, allowing it to dynamically adapt to the evolving information needs in multi-step reasoning.  

\section{Conclusion}
This paper proposes \OURS{} to enhance retrieval process of iRAG models. 
\OURS{} decomposes documents into knowledge triples and employs a knowledge-driven iterative retrieval framework to systematically retrieve relevant knowledge triples. 
The retrieved triples are used to rank documents, which serve as inputs for answer generation. 
Empirical results show that \OURS{} achieves significant retrieval and QA improvements, with an average increase of 9.40\% in R$\text{@}3$ and 5.14\% in F1, highlighting its effectiveness in multi-hop QA. 

\clearpage
\section*{Limitations}
We identify the following limitations of our work: 
(1) The Aligner model is trained using silver data constructed from only three multi-hop QA datasets. While our results demonstrate its effectiveness, we leave the exploration of methods to construct larger-scale and higher-quality training data for future work;
(2) In \OURS{}, we train only the Aligner model and keep the Constructor model frozen. While further training the Constructor could potential improve performance, we choose to keep it frozen to maintain our framework' adaptability to different LLMs, rather than relying on a specific fine-tuned LLM. 
Appendix~\ref{app:performance_different_llm} provides a detailed analysis of the performance using different LLM-based Constructor within our framework.

\bibliography{custom}

\appendix

\section{Prompts}

\subsection{Prompt for Knowledge Triple Extraction}
\label{app:prompt_knowledge_decomposition}

The prompt used for extracting knowledge triples from a document is illustrated in Figure~\ref{figure:prompt_knowledge_triple_extraction}. 

\begin{figure}[ht]
\centering
\begin{tcolorbox}[colback=gray!5!white,colframe=gray!60!black,title=Prompt Used for Knowledge Triple Extraction,left=0.2mm, right=0.2mm,bottom=0.2mm]
\begin{normalsize}
    \textbf{Instruction}: You are a knowledge graph constructor tasked with extracting knowledge triples in the form of <head entity; relation; tail entity> from a document. Each triple denotes a specific relationship between entities or an event. The head entity and tail entity can be the provided title or phrases in the text. If multiple tail entities share the same relation with a head entity, aggregate these tail entities using commas. Format your output in the form of <head entity; relation; tail entity>. 
    
    \vspace{0.5em} \textbf{Examples}: 
    
    {\fontsize{10}{4}\selectfont {Title}: \textit{Dana Blankstein}}
    
    {\fontsize{10}{4}\selectfont {Text}: \textit{Dana Blankstein- Cohen( born March 3, 1981) is the director of the Israeli Academy of Film and Television. She is a film director, and an Israeli culture entrepreneur.}}

    {\fontsize{10}{4}\selectfont {Knowledge Triples}: \textit{<Dana Blankstein; full name; Dana Blankstein-Cohen>, <Dana Blankstein; birth date; March 3, 1981>, <Dana Blankstein; nationality; Israeli>, <Dana Blankstein; position; director of the Israeli Academy of Film and Television>, <Dana Blankstein; profession; film director, culture entrepreneur>}}

    \vspace{0.5em} \textbf{Inputs}:

    {\fontsize{10}{4}\selectfont {Title}: \textit{\{\textit{document title}\}}}
    
    {\fontsize{10}{4}\selectfont {Text}: \textit{\{\textit{document text}\}}}

    {\fontsize{10}{4}\selectfont {Knowledge Triples}:}
    
\end{normalsize}
\end{tcolorbox}
\vspace{-1.0em}
\caption{Prompt used for extracting knowledge triples.}
\label{figure:prompt_knowledge_triple_extraction}
\vspace{-0.5em}
\end{figure}

\subsection{Prompt for Reasoning Chain Construction}
\label{app:prompt_reasoning_chain_construction}

The prompt used by the reasoning chain constructor to build reasoning chains is illustrated in Figure~\ref{figure:prompt_constructor}, where we instruct the Reasoning Chain Constructor to complete the $i$-th step reasoning chain with the provided candidate knowledge triples.  

\begin{figure}[ht]
\centering
\begin{tcolorbox}[colback=gray!5!white,colframe=gray!60!black,title=Prompt Used by Reasoning Chain Constructor,left=0.2mm, right=0.2mm,bottom=0.2mm]
\begin{normalsize}
    \textbf{Instruction}: Follow the examples to answer the input question by reasoning step-by-step. Output both reasoning steps and the answer.

    \textbf{Examples}: 
    
    {\fontsize{10}{4}\selectfont \textbf{Question}: \textit{Consider the racer for whom the bend at the 26th Milestone, Isle of Man is dedicated. When were they born?}}
    
    {\fontsize{10}{4}\selectfont \textbf{Thought}: \textit{<26th Milestone, Isle of Man; named after; Joey Dunlop>,<Joey Dunlop; date of birth; 25 February 1952>. So the answer is 25 February 1952.}}.... 

    \textbf{Inputs}:

    {\fontsize{10}{4}\selectfont \textbf{Context}: \{\textit{candidate triples}\}}
    
    {\fontsize{10}{4}\selectfont \textbf{Question}: \{\textit{question}\}}
    
    {\fontsize{10}{4}\selectfont \textbf{Thought}: \{\textit{$i$-th step reasoning chain}\} }
\end{normalsize}
\end{tcolorbox}
\vspace{-1.0em}
\caption{Prompt used by Reasoning Chain Constructor.}
\label{figure:prompt_constructor}
\vspace{-0.5em}
\end{figure}

\subsection{Prompt for Answer Generation}
\label{app:prompt_answer_generation}

The prompt used by the reader to generate answers is illustrated in Figure~\ref{figure:prompt_answer_generation}. 

\begin{figure}[ht]
\centering
\begin{tcolorbox}[colback=gray!5!white,colframe=gray!60!black,title=Prompt Used for Answer Generation,left=0.2mm, right=0.2mm,bottom=0.2mm]
\begin{normalsize}
    \textbf{Instruction}: Given some context and a question, please only output the answer to the question. 
    
    \textbf{Inputs}:

    {\fontsize{10}{4}\selectfont \textbf{Context}: \{\textit{retrieved documents}\}}
    
    {\fontsize{10}{4}\selectfont \textbf{Question}: \{\textit{question}\}}
    
    {\fontsize{10}{4}\selectfont \textbf{Answer}: }
\end{normalsize}
\end{tcolorbox}
\vspace{-1.0em}
\caption{Prompt used by the Reader.}
\label{figure:prompt_answer_generation}
\vspace{-1em}
\end{figure}

\section{Experimental Details}
\label{app:experimental_details}

\subsection{Datasets}
\label{app:datasets}
In our experiments, we employ five multi-hop QA datasets: HotPotQA, 2WikiMultiHopQA (2Wiki), MuSiQue, Bamboogle as well as  WebQuestions (WebQA), and one single-hop QA dataset: Natural Questions (NQ). 
For HotpotQA, we use the corpus provided by its authors for retrieval.  
For 2WikiMultihopQA and MuSiQue, we construct the retrieval corpus following the exact same procedure outlined by~\citet{trivedi2023interleaving}. For all other datasets, we leverage the Wikipedia corpus introduced by~\citet{karpukhin2020dense}. 

For datasets with public test sets (Bamboogle, WebQA and NQ), we report performance on their full test sets. 
For those with non-public test sets (HotPotQA, 2Wiki and MuSiQue), we use their full development sets as test sets and report the corresponding performance. Since these three datasets are also used for training, we randomly select 500 questions from their original training sets to serve as development sets, while the remaining questions are used for training. The statistics of experimental datasets can be found in Table~\ref{table:statistics}. 
Moreover, in \OURS{}, we precompute knowledge triples for all the documents in the corpus. The statistics of the resulting KG corpus are also provided in Table~\ref{table:statistics}. 

\begin{table*}[tb]
\centering
\resizebox{0.9\textwidth}{!}{
\begin{tabular}{lcccccccccccc}
\toprule
& \multicolumn{3}{c}{\textbf{HotPotQA}} & \multicolumn{3}{c}{\textbf{2Wiki}} & \multicolumn{3}{c}{\textbf{MuSiQue}} & \textbf{Bamboogle} & \textbf{WebQA} & \textbf{NQ} \\
\cmidrule(lr){2-4} \cmidrule(lr){5-7} \cmidrule(lr){8-10} \cmidrule(lr){11-11} \cmidrule(lr){12-12} \cmidrule(lr){13-13}
& Train & Dev. & Test & Train & Dev. & Test & Train & Dev. & Test & Test & Test & Test \\
\midrule
\noalign{\vskip 0.5ex} \multicolumn{7}{l}{\textbf{\textit{Statistics of Experimental Datasets}}} \vspace{0.3em} \\
\textbf{\# Questions} & 89,947 & 500 & 7,405 & 166,954 & 500 & 12,576 & 19,438 & 500 & 2,417 & 125 & 2,032 & 3,610 \\

\cmidrule(lr){1-13}
\noalign{\vskip 0.5ex} \multicolumn{7}{l}{\textbf{\textit{Statistics of Retrieval Corpus}}} \vspace{0.3em} \\
\textbf{Corpus} & \multicolumn{3}{c}{HotPotQA} & \multicolumn{3}{c}{2WikiMultiHopQA} & \multicolumn{3}{c}{MuSiQue} & Wikipedia & Wikipedia & Wikipedia \\
\textbf{\# Documents} & \multicolumn{3}{c}{5M} & \multicolumn{3}{c}{431K} & \multicolumn{3}{c}{117K} & 21M & 21M & 21M \\

\cmidrule(lr){1-13}\noalign{\vskip 0.5ex}
\multicolumn{7}{l}{\textit{\textbf{Statistics of the Extracted Knowledge Graph Corpus}}}\vspace{0.3em} \\ 
\textbf{Avg. \# Entities per Document} & \multicolumn{3}{c}{6.93} & \multicolumn{3}{c}{8.16} & \multicolumn{3}{c}{9.40} & 11.12 & 11.12 & 11.12 \\ 
\textbf{Avg. \# Triples per Document}  & \multicolumn{3}{c}{5.91} & \multicolumn{3}{c}{7.32} & \multicolumn{3}{c}{8.20} & \textcolor{white}{0}8.33 & \textcolor{white}{0}8.33 & \textcolor{white}{0}8.33 \\ 
\bottomrule    
\end{tabular}
}
\vspace{-0.5em}
\caption{Statistics of experimental datasets, retrieval corpus, and pre-computed knowledge graph corpus.} 
\vspace{-1.0em}
\label{table:statistics}
\end{table*}

\subsection{Baselines}
\label{app:baselines}

\textit{Standard RAG} model follows the vanilla retriever-reader pipeline, where the retriever model first retrieves top-$K$ documents from the corpus and the reader model then generates answers based on these retrieved documents. 
For \textit{IRCoT} and \textit{FLARE}, we use the implementations provided by FlashRAG~\cite{jin2024flashrag}. For other models, including \textit{DRAGIN}, \textit{BeamDR} and \textit{Vector-PRF}, we adapt the code released by their authors to align with our experimental setup. 
Notably, for fair comparison, both our \OURS{} and baselines use the same retriever for retrieving documents from the corpus and the same reader for generating answers. 

\subsection{Training and Hyperparameter Details}
\label{app:training_hparam_details}

\paragraph{Training Data Construction.}
We generate training data for the Reasoning Chain Aligner using existing multi-hop QA datasets. Specifically, for each multi-hop question and its ground-truth relevant documents, we apply TRACE~\cite{fang2024trace} (using the default hyperparameter setting) to construct five potential knowledge triple-based reasoning chains for answering the question. For each chain, we use Llama3 as the reader to generate an answer based on the context provided by the chain. The first chain that successfully produces the correct answer is selected as the ground-truth reasoning chain for that question. The question and its ground-truth reasoning chain will serve as labeled data for training.  
We filter out questions where all reasoning chains fail to produce the correct answer. 
In practice, we build training data from the \textit{training sets} of three multi-hop QA datasets: HotPotQA, 2Wiki and MuSiQue. In addition, we use the same procedure to construct development data from the development sets of these three datasets for hyperparameter tuning. 
The statistics of the data used to train the Aligner are presented in Table~\ref{table:statistics_of_training_data}. 

\begin{table}[tb]
\centering
\resizebox{0.35\textwidth}{!}{
\begin{tabular}{p{4cm}cc}
\toprule
& \textbf{Train}  & \textbf{Dev.} \\
\midrule
\textbf{\# Questions} & 115,567 & 815 \\
\textbf{Avg. Chain Length} & 2.36 & 2.35\\
\bottomrule
\end{tabular}
} 
\vspace{-0.5em}
\caption{Statistics of the data used for training the Reasoning Chain Aligner.}
\label{table:statistics_of_training_data}
\vspace{-1.0em}
\end{table}

\vspace{0.5em} \noindent \textbf{Training Details.}
For an incomplete reasoning chain $r$, we treat the correct next triple as positive sample. 
To generate negative samples, we follow the procedure described in ``Knowledge Decomposition'' section to obtain a set of candidate triples $\tilde{\mathcal{T}^i}$. 
The training process uses the Adam optimizer~\cite{kingma2015adam} with a learning rate of 2e-5 and a weight decay of 0.01. We set the batch size to 64, include 7 negative samples per data point, and use a temperature parameter $\tau$ of 0.01. The Aligner is trained for 10 epochs, and we select the checkpoint with the best performance (R$\text{@}$5) on the development set.

\begin{table}[tb]
\centering
\resizebox{0.45\textwidth}{!}{
\begin{tabular}{cc}
\toprule
\textbf{Model} & \textbf{Huggingface Checkpoint} \\
\midrule
\textbf{E5} & {\tt intfloat/e5-large-v2} \\
\textbf{BGE} & {\tt BAAI/bge-large-en-v1.5} \\
\textbf{Llama3} & {\tt meta-llama/Meta-Llama-3-8B-Instruct} \\
\textbf{Mistral} & {\tt mistralai/Mistral-7B-Instruct-v0.2} \\
\textbf{Gemma2} & {\tt google/gemma-2-9b-it} \\
\textbf{Qwen2.5} & {\tt Qwen/Qwen2.5-7B-Instruct}\\
\textbf{Flan-T5} & {\tt google/flan-t5-xl} \\
\bottomrule
\end{tabular}
} 
\vspace{-0.5em}
\caption{The specific huggingface checkpoints used in our experiments.}
\label{table:specific_model_used_in_kirag}
\vspace{-1.0em}
\end{table}

\vspace{0.5em} \noindent \textbf{Implementation and Hyperparameter Details.}
Throughout the experiments, we set the maximum number of iterative steps $L$ to 5. The details of each component in our \OURS{} are outlined as follows:

For the \textit{Retriever} model, we use either E5~\cite{wang2022text} or BGE~\cite{xiao2023c} to retrieve documents. The number of retrieved documents per iteration (i.e., $K_0$) is $10$. For the \textit{Knowledge Decomposition} component, we use Llama3~\cite{dubey2024llama} to extract knowledge triples for each retrieved document. For the \textit{Reasoning Chain Aligner}, given the question and partial reasoning chain, it selects top-20 (i.e., $N=20$) knowledge triples that are likely to extend the existing chain. 

For the \textit{Reasoning Chain Constructor}, we try different LLMs, including Llama3, Mistral~\cite{jiang2023mistral} and Gemma2~\cite{team2024gemma}, to select a triple to extend the partial reasoning chain for subsequent retrieval. We main report the performance of using Llama3 as the Constructor as it achieves the best performance (see Appendix~\ref{app:performance_different_llm}). \jy{Moreover, when completing the partial reasoning chain, we filter triples that are not present in the provided candidate set to ensure factual reliability.} 

Moreover, we leverage different readers to evaluate the QA performance, which includes Llama3, Qwen2.5~\cite{yang2024qwen2}, Flan-T5~\cite{chung2024scaling} and TRACE~\cite{fang2024trace}.  
The specific huggingface checkpoints we used in our experiments are provided in Table~\ref{table:specific_model_used_in_kirag}. 

\section{Additional Experimental Results and Analysis}
\label{app:additional_results_and_analysis}

\begin{table}[tb]
\centering
\resizebox{0.48\textwidth}{!}{
\begin{tabular}{lcccccc}
\toprule
\multirow{2}{*}{\textbf{Model}} & \multicolumn{2}{c}{\textbf{HotPotQA}} & \multicolumn{2}{c}{\textbf{2Wiki}} & \multicolumn{2}{c}{\textbf{MuSiQue}}\\
\cmidrule(lr){2-3} \cmidrule(lr){4-5} \cmidrule(lr){6-7}
& \textbf{R}$\text{@}$\textbf{3} & \textbf{R}$\text{@}$\textbf{5} & \textbf{R}$\text{@}$\textbf{3} & \textbf{R}$\text{@}$\textbf{5} & \textbf{R}$\text{@}$\textbf{3} & \textbf{R}$\text{@}$\textbf{5} \\
\midrule
\textbf{RAG} & 64.46\textcolor{white}{$^*$} & 69.71\textcolor{white}{$^*$} & 60.50\textcolor{white}{$^*$} & 64.91\textcolor{white}{$^*$} & 39.40\textcolor{white}{$^*$} & 45.16\textcolor{white}{$^*$}  \\
\hdashline\noalign{\vskip 0.5ex}
\textbf{Vector-PRF} & 64.38\textcolor{white}{$^*$} & 69.34\textcolor{white}{$^*$} & 60.19\textcolor{white}{$^*$} & 64.37\textcolor{white}{$^*$} &39.16\textcolor{white}{$^*$} & 43.86\textcolor{white}{$^*$} \\
\hdashline\noalign{\vskip 0.5ex}
\textbf{FLARE} & 53.63\textcolor{white}{$^*$} & 58.83\textcolor{white}{$^*$} & 60.28\textcolor{white}{$^*$} & 69.30\textcolor{white}{$^*$} & 37.10\textcolor{white}{$^*$} & 43.16\textcolor{white}{$^*$} \\
\textbf{DRAGIN} & \underline{71.71}\textcolor{white}{$^*$} & \underline{76.93}\textcolor{white}{$^*$} & \underline{62.42}\textcolor{white}{$^*$} & 71.14\textcolor{white}{$^*$} & \underline{45.78}\textcolor{white}{$^*$} & \underline{52.44}\textcolor{white}{$^*$} \\
\textbf{IRCoT} & 69.96\textcolor{white}{$^*$} & 75.62\textcolor{white}{$^*$} & 60.20\textcolor{white}{$^*$} & \underline{72.23}\textcolor{white}{$^*$} & 42.13\textcolor{white}{$^*$} & 48.91\textcolor{white}{$^*$} \\
\hdashline\noalign{\vskip 0.5ex}
\textbf{\OURS{}-Doc} & 52.42\textcolor{white}{$^*$} & 67.81\textcolor{white}{$^*$} & 41.42\textcolor{white}{$^*$} & 56.55\textcolor{white}{$^*$} & 28.64\textcolor{white}{$^*$} & 39.82\textcolor{white}{$^*$} \\
\textbf{\OURS{}-Sent} & 47.81\textcolor{white}{$^*$} & 62.99\textcolor{white}{$^*$} & 41.42\textcolor{white}{$^*$} & 56.55\textcolor{white}{$^*$} & 28.27\textcolor{white}{$^*$} & 38.26\textcolor{white}{$^*$}  \\
\textbf{\OURS{}} & \textbf{79.69}$^\dagger$ & \textbf{83.61}$^\dagger$ & \textbf{78.50}$^\dagger$ & \textbf{88.94}$^\dagger$ & \textbf{52.62}$^\dagger$ & \textbf{58.39}$^\dagger$ \\
\bottomrule    
\end{tabular}
}
\vspace{-0.5em}
\caption{Retrieval performance (\%) using BGE as the retriever model, where the best and the second-best results are marked in bold and underlined, respectively, and $\dagger$ denotes p-value<0.05 compared to the best-performing baseline. Results for BeamDR are omitted as it relies on its own trained BERT model for retrieval, yielding the same results as presented in Table~\ref{table:in_domain_retrieval_performance}.}
\vspace{-0.5em}
\label{table:retrieval_performance_bge}
\end{table}

\begin{table}[tb]
\centering
\resizebox{0.48\textwidth}{!}{
\begin{tabular}{lcccccc}
\toprule
\multirow{2}{*}{\textbf{Model}} & \multicolumn{2}{c}{\textbf{HotPotQA}} & \multicolumn{2}{c}{\textbf{2Wiki}} & \multicolumn{2}{c}{\textbf{MuSiQue}}\\
\cmidrule(lr){2-3} \cmidrule(lr){4-5} \cmidrule(lr){6-7}
& \textbf{EM} & \textbf{F1} & \textbf{EM} & \textbf{F1} & \textbf{EM} & \textbf{F1} \\
\midrule
\textbf{RAG} & 34.11\textcolor{white}{$^*$} & 46.68\textcolor{white}{$^*$} & 14.73\textcolor{white}{$^*$} & 30.16\textcolor{white}{$^*$} & \textcolor{white}{0}8.77\textcolor{white}{$^*$} & 17.01\textcolor{white}{$^*$} \\
\hdashline\noalign{\vskip 0.5ex}
\textbf{Vector-PRF} & 33.94\textcolor{white}{$^*$} & 46.58\textcolor{white}{$^*$} & 14.70\textcolor{white}{$^*$} & 30.00\textcolor{white}{$^*$} & \textcolor{white}{0}8.65\textcolor{white}{$^*$} & 16.97\textcolor{white}{$^*$} \\
\hdashline\noalign{\vskip 0.5ex}
\textbf{BeamDR} & 38.34\textcolor{white}{$^*$} & 51.64\textcolor{white}{$^*$} & 14.42\textcolor{white}{$^*$} & 27.25\textcolor{white}{$^*$} & \textcolor{white}{0}7.08\textcolor{white}{$^*$} & 14.42\textcolor{white}{$^*$}  \\ 
\textbf{FLARE} & 34.49\textcolor{white}{$^*$} & 46.65\textcolor{white}{$^*$} & \underline{25.01}\textcolor{white}{$^*$} & 40.59\textcolor{white}{$^*$} & 13.07\textcolor{white}{$^*$} & 21.38\textcolor{white}{$^*$} \\
\textbf{DRAGIN} & 41.73\textcolor{white}{$^*$} & 55.68\textcolor{white}{$^*$} & 24.62\textcolor{white}{$^*$} & \underline{40.69}\textcolor{white}{$^*$} & \underline{16.43}\textcolor{white}{$^*$} & \underline{26.29}\textcolor{white}{$^*$} \\
\textbf{IRCoT} & \underline{43.18}\textcolor{white}{$^*$} & \underline{57.08}\textcolor{white}{$^*$} & 24.25\textcolor{white}{$^*$} & 40.12\textcolor{white}{$^*$} & 14.89\textcolor{white}{$^*$} & 23.99\textcolor{white}{$^*$} \\
\hdashline\noalign{\vskip 0.5ex}
\textbf{\OURS{}-Doc} & 30.47\textcolor{white}{$^*$} & 42.52\textcolor{white}{$^*$} & 11.97\textcolor{white}{$^*$} & 23.97\textcolor{white}{$^*$} & \textcolor{white}{0}7.03\textcolor{white}{$^*$} & 14.59\textcolor{white}{$^*$} \\
\textbf{\OURS{}-Sent} & 30.44\textcolor{white}{$^*$} & 42.00\textcolor{white}{$^*$} & 12.82\textcolor{white}{$^*$} & 25.22\textcolor{white}{$^*$} & \textcolor{white}{0}8.56\textcolor{white}{$^*$} & 16.16\textcolor{white}{$^*$} \\
\textbf{\OURS{}} & \textbf{45.16}$^\dagger$ & \textbf{59.85}$^\dagger$ & \textbf{35.02}$^\dagger$ & \textbf{54.01}$^\dagger$ & \textbf{18.87}$^\dagger$ & \textbf{29.17}$^\dagger$ \\
\bottomrule    
\end{tabular}
} 
\vspace{-0.5em}
\caption{QA performance (\%) using BGE as the Retriever. The best and second-best performance are highlighted in bold and underlined, respectively. $\dagger$ indicates p-value<0.05 compared with best-performing baseline.}
\vspace{-0.5em}
\label{table:qa_performance_bge}
\end{table}

\begin{table}[tb]
\centering
\resizebox{0.48\textwidth}{!}{
\begin{tabular}{llcccccc}
\toprule
\multirow{2}{*}{\textbf{Reader}}& \multirow{2}{*}{\textbf{Model}} & \multicolumn{2}{c}{\textbf{HotPotQA}} & \multicolumn{2}{c}{\textbf{2Wiki}} & \multicolumn{2}{c}{\textbf{MuSiQue}}\\
\cmidrule(lr){3-4} \cmidrule(lr){5-6} \cmidrule(lr){7-8}
& & \textbf{EM} & \textbf{F1} & \textbf{EM} & \textbf{F1} & \textbf{EM} & \textbf{F1} \\
\midrule
\multirow{8}{*}{\textbf{Qwen2.5}}
& \textbf{RAG} 
& 34.69\textcolor{white}{$^*$} & 46.15\textcolor{white}{$^*$} & 33.37\textcolor{white}{$^*$} & 38.51\textcolor{white}{$^*$} & \textcolor{white}{0}9.14\textcolor{white}{$^*$} & 17.17\textcolor{white}{$^*$} \\
\cdashline{2-8}
& \textbf{Vector-PRF} 
& 34.54\textcolor{white}{$^*$} & 46.12\textcolor{white}{$^*$} & 33.41\textcolor{white}{$^*$} & 38.51\textcolor{white}{$^*$} & \textcolor{white}{0}8.90\textcolor{white}{$^*$} & 16.91\textcolor{white}{$^*$} \\
& \textbf{BeamDR} 
& 39.61\textcolor{white}{$^*$} & 51.51\textcolor{white}{$^*$} & 22.41\textcolor{white}{$^*$} & 29.95\textcolor{white}{$^*$} & \textcolor{white}{0}6.70\textcolor{white}{$^*$} & 14.06\textcolor{white}{$^*$}  \\ 
\cdashline{2-8}
& \textbf{FLARE} 
& 36.30\textcolor{white}{$^*$} & 47.33\textcolor{white}{$^*$} & 36.73\textcolor{white}{$^*$} & 44.38\textcolor{white}{$^*$} & 12.58\textcolor{white}{$^*$} & 21.22\textcolor{white}{$^*$}  \\
& \textbf{DRAGIN} 
& 44.07\textcolor{white}{$^*$} & 56.91\textcolor{white}{$^*$} & 36.75\textcolor{white}{$^*$} & 44.49\textcolor{white}{$^*$} & 18.16\textcolor{white}{$^*$} & 28.68\textcolor{white}{$^*$} \\
& \textbf{IRCoT} 
& 43.44\textcolor{white}{$^*$} & 56.46\textcolor{white}{$^*$} & 38.39\textcolor{white}{$^*$} & 45.97\textcolor{white}{$^*$} & 15.60\textcolor{white}{$^*$} & 25.36\textcolor{white}{$^*$} \\
\cdashline{2-8}
& \textbf{\OURS{}-Doc} 
& 35.68\textcolor{white}{$^*$} & 47.15\textcolor{white}{$^*$} & 29.23\textcolor{white}{$^*$} & 34.51\textcolor{white}{$^*$} & \textcolor{white}{0}7.61\textcolor{white}{$^*$} & 15.27\textcolor{white}{$^*$} \\
& \textbf{\OURS{}-Sent}
& 34.21\textcolor{white}{$^*$} & 45.64\textcolor{white}{$^*$} & 29.78\textcolor{white}{$^*$} & 34.92\textcolor{white}{$^*$} & \textcolor{white}{0}9.64\textcolor{white}{$^*$} & 17.83\textcolor{white}{$^*$} \\
& \textbf{\OURS{}} & \textbf{47.89}{$^\dagger$} & \textbf{61.41}$^\dagger$ & \textbf{47.42}$^\dagger$ & \textbf{56.02}{$^\dagger$} & \textbf{19.73}{$^\dagger$} & \textbf{30.79}{$^\dagger$} \\
\midrule
\multirow{8}{*}{\textbf{Flan-T5}}
& \textbf{RAG}
& 37.08\textcolor{white}{$^*$} & 47.32\textcolor{white}{$^*$} & 17.42\textcolor{white}{$^*$} & 22.05\textcolor{white}{$^*$} & \textcolor{white}{0}8.94\textcolor{white}{$^*$} & 15.06\textcolor{white}{$^*$} \\
\cdashline{2-8}
& \textbf{Vector-PRF} 
& 37.02\textcolor{white}{$^*$} & 47.23\textcolor{white}{$^*$} & 31.58\textcolor{white}{$^*$} & 36.26\textcolor{white}{$^*$} & \textcolor{white}{0}8.98\textcolor{white}{$^*$} & 15.16\textcolor{white}{$^*$} \\
& \textbf{BeamDR} 
& 41.89\textcolor{white}{$^*$} & 52.83\textcolor{white}{$^*$} & 18.73\textcolor{white}{$^*$} & 23.29\textcolor{white}{$^*$} & \textcolor{white}{0}7.03\textcolor{white}{$^*$} & 12.21\textcolor{white}{$^*$}   \\ 
\cdashline{2-8}
& \textbf{FLARE} 
& 39.81\textcolor{white}{$^*$} & 50.46\textcolor{white}{$^*$} & 33.31\textcolor{white}{$^*$} & 39.95\textcolor{white}{$^*$} & 13.28\textcolor{white}{$^*$} & 19.74\textcolor{white}{$^*$}  \\
& \textbf{DRAGIN} 
& 46.75\textcolor{white}{$^*$} & 58.52\textcolor{white}{$^*$} & 34.19\textcolor{white}{$^*$} & 40.68\textcolor{white}{$^*$} & 18.12\textcolor{white}{$^*$} & 25.22\textcolor{white}{$^*$}  \\
& \textbf{IRCoT} 
& {47.32}\textcolor{white}{$^*$} & 59.05\textcolor{white}{$^*$} & 35.90\textcolor{white}{$^*$} & 42.31\textcolor{white}{$^*$} & 16.42\textcolor{white}{$^*$} & 23.61\textcolor{white}{$^*$}  \\
\cdashline{2-8}
& \textbf{\OURS{}-Doc}
& 36.18\textcolor{white}{$^*$} & 46.45\textcolor{white}{$^*$} & 25.03\textcolor{white}{$^*$} & 29.58\textcolor{white}{$^*$} & \textcolor{white}{0}7.45\textcolor{white}{$^*$} & 13.53\textcolor{white}{$^*$} \\
& \textbf{\OURS{}-Sent}
& 38.86\textcolor{white}{$^*$} & 49.59\textcolor{white}{$^*$} & 31.27\textcolor{white}{$^*$} & 36.58\textcolor{white}{$^*$} & 11.34\textcolor{white}{$^*$} & 17.98\textcolor{white}{$^*$} \\
& \textbf{\OURS{}} & \textbf{49.31}{$^\dagger$} & \textbf{61.38}{$^\dagger$} & \textbf{39.99}{$^\dagger$} & \textbf{46.51}{$^\dagger$} & \textbf{19.07}\textcolor{white}{$^\dagger$} & \textbf{27.29}{$^\dagger$} \\
\midrule
\multirow{8}{*}{\textbf{TRACE}}
& \textbf{RAG} &39.18\textcolor{white}{$^\dagger$} & 51.82\textcolor{white}{$^\dagger$} & 21.10\textcolor{white}{$^\dagger$} & 34.28\textcolor{white}{$^\dagger$} & 11.63\textcolor{white}{$^*$} & 19.49\textcolor{white}{$^*$} \\
\cdashline{2-8}
& \textbf{Vector-PRF} & 38.85\textcolor{white}{$^*$} & 51.54\textcolor{white}{$^*$} & 21.91\textcolor{white}{$^*$} &34.81\textcolor{white}{$^*$} & 11.58\textcolor{white}{$^*$} & 19.59\textcolor{white}{$^*$} \\
& \textbf{BeamDR} &43.21\textcolor{white}{$^*$} &56.28\textcolor{white}{$^*$} & 21.01\textcolor{white}{$^*$} & 33.23\textcolor{white}{$^*$} & 10.67\textcolor{white}{$^*$} & 18.26\textcolor{white}{$^*$} \\ 
\cdashline{2-8}
& \textbf{FLARE} & 39.31\textcolor{white}{$^\dagger$} & 51.40\textcolor{white}{$^\dagger$} & 31.85\textcolor{white}{$^*$} & 45.50\textcolor{white}{$^*$} & 14.89\textcolor{white}{$^*$} & 23.75\textcolor{white}{$^*$} \\
& \textbf{DRAGIN} & 44.29\textcolor{white}{$^*$} & 57.64\textcolor{white}{$^*$} & 31.88\textcolor{white}{$^*$} & 45.55\textcolor{white}{$^*$} & 18.11\textcolor{white}{$^*$} & 27.41\textcolor{white}{$^*$} \\
& \textbf{IRCoT} & 45.29\textcolor{white}{$^*$} & 58.77\textcolor{white}{$^*$} & 32.05\textcolor{white}{$^*$} & 46.55\textcolor{white}{$^*$}  & 16.84\textcolor{white}{$^*$} & 25.78\textcolor{white}{$^*$} \\
\cdashline{2-8}
& \textbf{\OURS{}-Doc} & 45.36\textcolor{white}{$^*$} & 58.82\textcolor{white}{$^*$} & 29.41\textcolor{white}{$^*$} & 43.99\textcolor{white}{$^*$} & 13.69\textcolor{white}{$^*$} & 22.44\textcolor{white}{$^*$} \\
& \textbf{\OURS{}-Sent} &43.38\textcolor{white}{$^*$} & 56.68\textcolor{white}{$^*$} & 27.51\textcolor{white}{$^*$} & 41.92\textcolor{white}{$^*$} & 16.84\textcolor{white}{$^*$} & 25.59\textcolor{white}{$^*$} \\
& \textbf{\OURS{}} &\textbf{46.41}\textcolor{white}{$^\dagger$} &\textbf{60.22}{$^\dagger$} & \textbf{33.13}\textcolor{white}{$^\dagger$} & \textbf{48.49}{$^\dagger$} & \textbf{19.32}\textcolor{white}{$^\dagger$} & \textbf{29.10}{$^\dagger$} \\
\bottomrule
\end{tabular}
}
\vspace{-0.5em}
\caption{QA performance (\%) using different Reader models, where $\dagger$ indicates p-value<0.05 compared with best-performing baseline.}
\label{table:qa_performance_e5_different_readers}
\end{table}

\subsection{Overall Performance of Using Different Retrievers and Readers}
\label{app:overall_performance_different_retrievers_readers}

To validate the effectiveness of \OURS{}, we provide additional results using different retrievers and readers. 
Specifically, we replace the E5 Retriever with BGE Retriever for retrieving documents from the corpus and the other components remain unchanged. The corresponding retrieval and QA performance are presented in Table~\ref{table:retrieval_performance_bge} and Table~\ref{table:qa_performance_bge}, respectively. 
The results are consistent with those obtained using the E5 Retriever, demonstrating the adaptability and effectiveness of our \OURS{} across different retriever models. 

Moreover, to assess the quality of the documents retrieved by \OURS{}, we report QA performance using different reader models in Table~\ref{table:qa_performance_e5_different_readers}. The results suggest that \OURS{} consistently outperforms all the baselines across different readers, demonstrating its ability to provide high-quality retrieval results that enhance downstream QA performance.  

\begin{table}[tb]
\centering
\resizebox{0.48\textwidth}{!}{
\begin{tabular}{lcccccc}
\toprule
\multirow{2}{*}{\textbf{Model}} & \multicolumn{2}{c}{\textbf{HotPotQA}} & \multicolumn{2}{c}{\textbf{2Wiki}} & \multicolumn{2}{c}{\textbf{MuSiQue}}\\
\cmidrule(lr){2-3} \cmidrule(lr){4-5} \cmidrule(lr){6-7}
& \textbf{EM} & \textbf{F1} & \textbf{EM} & \textbf{F1} & \textbf{EM} & \textbf{F1} \\
\midrule

\textbf{RAG} 
& 34.34\textcolor{white}{$^*$} & 47.72\textcolor{white}{$^*$} & 12.95\textcolor{white}{$^*$} & 29.94\textcolor{white}{$^*$} & \textcolor{white}{0}9.43\textcolor{white}{$^*$} & 17.50\textcolor{white}{$^*$} \\
\hdashline\noalign{\vskip 0.5ex}

\textbf{Vector-PRF} 
& 34.56\textcolor{white}{$^*$} & 47.87\textcolor{white}{$^*$} & 13.76\textcolor{white}{$^*$} & 30.42\textcolor{white}{$^*$} & 10.10\textcolor{white}{$^*$} & 17.74\textcolor{white}{$^*$} \\

\textbf{BeamDR} 
& 38.60\textcolor{white}{$^*$} & 52.24\textcolor{white}{$^*$} & 14.49\textcolor{white}{$^*$} & 28.71\textcolor{white}{$^*$} & \textcolor{white}{0}7.74\textcolor{white}{$^*$} & 15.13\textcolor{white}{$^*$}  \\ 
\hdashline\noalign{\vskip 0.5ex}

\textbf{FLARE} 
& 35.08\textcolor{white}{$^*$} & 48.07\textcolor{white}{$^*$} &  24.87\textcolor{white}{$^*$} &  41.95\textcolor{white}{$^*$} &  13.20\textcolor{white}{$^*$} &  22.23\textcolor{white}{$^*$} \\

\textbf{DRAGIN} 
& 41.16\textcolor{white}{$^*$} & 55.48\textcolor{white}{$^*$} &  24.47\textcolor{white}{$^*$} &  41.37\textcolor{white}{$^*$} & 17.54\textcolor{white}{$^*$} & 27.74\textcolor{white}{$^*$} \\

\textbf{IRCoT} 
& \underline{41.61}\textcolor{white}{$^*$} & \underline{56.01}\textcolor{white}{$^*$} & \underline{24.89}\textcolor{white}{$^*$} & \underline{42.90}\textcolor{white}{$^*$} & 14.85\textcolor{white}{$^*$} & 24.48\textcolor{white}{$^*$} \\
\hdashline\noalign{\vskip 0.5ex}

\textbf{\OURS{}-Doc} 
& 36.87\textcolor{white}{$^*$} & 50.68\textcolor{white}{$^*$} & 15.54\textcolor{white}{$^*$} & 32.17\textcolor{white}{$^*$} & \textcolor{white}{0}9.23\textcolor{white}{$^*$} & 17.50\textcolor{white}{$^*$} \\

\textbf{\OURS{}-Sent}
& 36.58\textcolor{white}{$^*$} & 50.09\textcolor{white}{$^*$} & 15.49\textcolor{white}{$^*$} & 31.59\textcolor{white}{$^*$} & 12.16\textcolor{white}{$^*$} & 20.66\textcolor{white}{$^*$}  \\

\textbf{\OURS{}} 
&\textbf{43.81}\textcolor{black}{$^\dagger$} & \textbf{58.42}$^\dagger$ & \textbf{27.26}$^\dagger$ & \textbf{47.59}$^\dagger$ & \textbf{17.58} & \textbf{28.92}$^\dagger$\\
\bottomrule    
\end{tabular}
}
\vspace{-0.5em}
\caption{QA performance (\%) using top-5 retrieved document as context. The best and second-best results marked in bold and underlined, respectively. \hspace{-0.2em}$^\dagger$ denote p-value<0.05 compared with best-performing baseline. }
\vspace{-0.7em}
\label{table:qa_performance_top5_document}
\end{table}

\subsection{QA Performance based on Top-5 Documents}
\label{app:qa_performance_top_5_docs}

Table~\ref{table:qa_performance_top5_document} presents the QA performance using the top-5 retrieved documents as the context, demonstrating similar results to those obtained with the top-3 retrieved documents.

\begin{table}[tb]
\centering
\resizebox{0.48\textwidth}{!}{
\begin{tabular}{lcccccc}
\toprule
\multirow{2}{*}{\textbf{Model}} & \multicolumn{2}{c}{\textbf{Bamboogle}} & \multicolumn{2}{c}{\textbf{WebQA}} & \multicolumn{2}{c}{\textbf{NQ}}\\
\cmidrule(lr){2-3} \cmidrule(lr){4-5} \cmidrule(lr){6-7}
& \textbf{EM} & \textbf{F1} & \textbf{EM} & \textbf{F1} & \textbf{EM} & \textbf{F1} \\
\midrule
\textbf{RAG} 
& 15.20\textcolor{white}{$^*$} & 22.66\textcolor{white}{$^*$} & 18.41\textcolor{white}{$^*$} & 31.04\textcolor{white}{$^*$} & 35.93\textcolor{white}{$^*$} & 41.18\textcolor{white}{$^*$}  \\
\hdashline\noalign{\vskip 0.5ex}

\textbf{Vector-PRF} 
& 15.20\textcolor{white}{$^*$} & 23.73\textcolor{white}{$^*$} & 18.31\textcolor{white}{$^*$} & 31.02\textcolor{white}{$^*$} & 36.09\textcolor{white}{$^*$} & 41.25\textcolor{white}{$^*$} \\

\textbf{BeamDR} & 11.20\textcolor{white}{$^*$} & 15.29\textcolor{white}{$^*$} & 15.50\textcolor{white}{$^*$} & 25.46\textcolor{white}{$^*$} & 21.63\textcolor{white}{$^*$} & 25.31\textcolor{white}{$^*$} \\ 
\hdashline\noalign{\vskip 0.5ex}

\textbf{FLARE} & 24.00\textcolor{white}{$^*$} & 31.93\textcolor{white}{$^*$} & \underline{20.57}\textcolor{white}{$^*$} & 31.47\textcolor{white}{$^*$} & 31.22\textcolor{white}{$^*$} & 35.17\textcolor{white}{$^*$} \\
\textbf{DRAGIN} 
& \underline{26.20}\textcolor{white}{$^*$} & \underline{37.68}\textcolor{white}{$^*$} & 20.37\textcolor{white}{$^*$} & \underline{32.31}\textcolor{white}{$^*$} & 35.43\textcolor{white}{$^*$} & 39.87\textcolor{white}{$^*$} \\
\textbf{IRCoT} 
& 21.60\textcolor{white}{$^*$} & 33.69\textcolor{white}{$^*$} & 19.39\textcolor{white}{$^*$} & 31.31\textcolor{white}{$^*$} & \textbf{37.34}\textcolor{white}{$^*$} & \textbf{42.50}\textcolor{white}{$^*$} \\
\hdashline\noalign{\vskip 0.5ex}

\textbf{\OURS{}-Doc}
& 17.60\textcolor{white}{$^*$} & 27.92\textcolor{white}{$^*$} & 18.36\textcolor{white}{$^*$} & 30.75\textcolor{white}{$^*$} & 33.63\textcolor{white}{$^*$} & 39.22\textcolor{white}{$^*$}  \\ 

\textbf{\OURS{}-Sent}
& 16.00\textcolor{white}{$^*$} & 28.15\textcolor{white}{$^*$} & 19.14\textcolor{white}{$^*$} & 31.27\textcolor{white}{$^*$} & 33.60\textcolor{white}{$^*$} & 38.12\textcolor{white}{$^*$}  \\

\textbf{\OURS{}}
& \textbf{29.60}$^\dagger$ & \textbf{42.00}$^\dagger$ & \textbf{20.67}\textcolor{white}{$^*$} & \textbf{32.87}\textcolor{white}{$^*$} & 36.29\textcolor{white}{$^*$} & 41.49\textcolor{white}{$^*$} \\
\bottomrule    
\end{tabular}
} 
\vspace{-0.5em}
\caption{QA performance (\%) on unseen multi-hop and single-hop QA datasets, where $^\dagger$ denotes p-value<0.05 compared with best-performing baselines.}
\vspace{-0.5em}
\label{table:unseen_qa_performance}
\end{table}

\begin{figure}[tb]
\begin{center}
\includegraphics[width=0.48\textwidth]{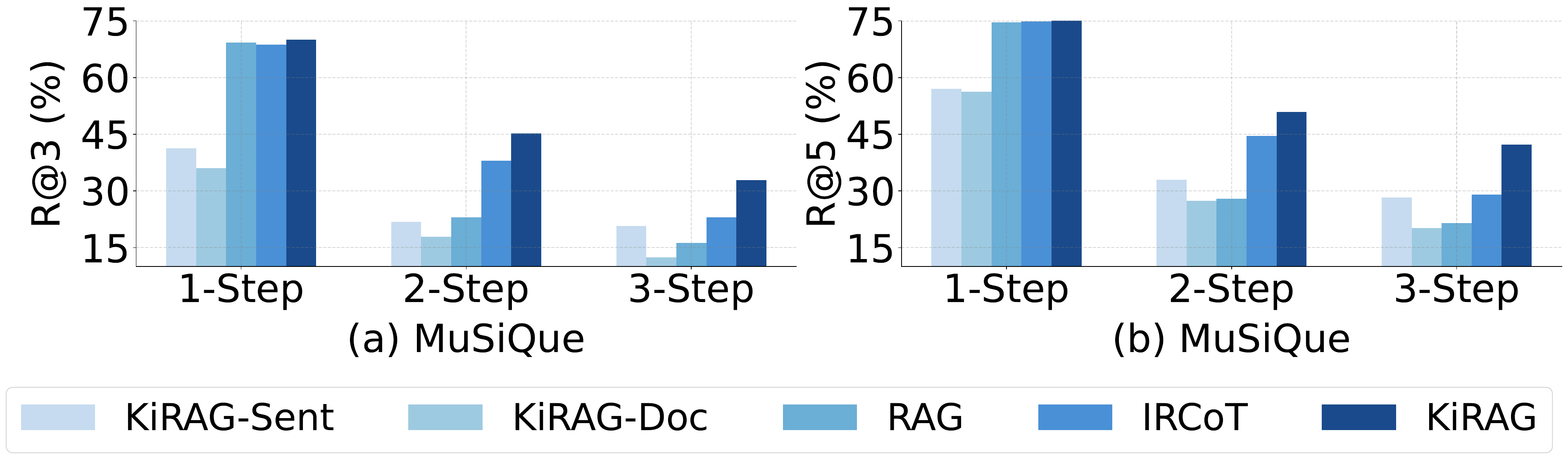}
\end{center}
\vspace{-0.5em}
\caption{Retrieval performance (\%) at different steps on MuSiQue dataset.}
\label{figure:op_recall_different_hop_musique}
\vspace{-0.5em}
\end{figure}

\subsection{Retrieval Performance at Different Steps on MuSiQue Dataset}
\label{app:retrieval_performance_different_step_musique}
Figure~\ref{figure:op_recall_different_hop_musique} presents the retrieval performance of \OURS{} and baseline methods at different steps on the MuSiQue dataset, showing similar trends to those observed on the HotPotQA and 2Wiki datasets.  

\subsection{QA Performance on Unseen Datasets}
\label{app:qa_performance_unseen_datasets}

Due to page limit, we present the QA performance on unseen multi-hop and single-hop QA datasets in Table~\ref{table:unseen_qa_performance}, which aligns with the retrieval performance reported in Table~\ref{table:unseen_retrieval_performance}. 
The results highlight that \OURS{} can effectively generalise to different types of QA tasks, maintaining high performance without overfitting to specific training data.

\begin{table*}[tb]
\centering
\begin{small}
\begin{tabular}{p{0.9\textwidth}}
\toprule
\textbf{Question}: Which film came out first, Blind Shaft or The Mask Of Fu Manchu? \vspace{0.3em} \\ 
\textbf{Relevant Knowledge Triples}: <Blind Shaft; release year; 2003>, <The Mask of Fu Manchu; release year; 1932> \\

\noalign{\vskip 1.0ex}\hdashline\noalign{\vskip 1.0ex}
\textbf{Question}: When did John V, Prince Of Anhalt-Zerbst's father die? \vspace{0.3em} \\ 
\textbf{Relevant Knowledge Triples}: <John V, Prince of Anhalt-Zerbst; father; Ernest I, Prince of Anhalt-Dessau>, <Ernest I, Prince of Anhalt-Dessau; death date; 12 June 1516> \\

\noalign{\vskip 1.0ex}\hdashline\noalign{\vskip 1.0ex}
\textbf{Question}: Which film has the director died first, Crimen A Las Tres or The Working Class Goes To Heaven? \vspace{0.3em} \\ 
\textbf{Relevant Knowledge Triples}: <Crimen a las tres; director; Luis Saslavsky>, <The Working Class Goes to Heaven; director; Elio Petri>, <Luis Saslavsky; death date; March 20, 1995>, <Elio Petri; death date; 10 November 1982> \\

\noalign{\vskip 1.0ex}\hdashline\noalign{\vskip 1.0ex}
\textbf{Question}: Who died first, Fleetwood Sheppard or George William Whitaker? \vspace{0.3em} \\ 
\textbf{Relevant Knowledge Triples}: <Fleetwood Sheppard; death date; 25 August 1698>, <George William Whitaker; death date; March 6, 1916> \\

\noalign{\vskip 1.0ex}\hdashline\noalign{\vskip 1.0ex}
\textbf{Question}: Who is the spouse of the director of film Eden And After? \vspace{0.3em} \\ 
\textbf{Relevant Knowledge Triples}: <Eden and After; director; Alain Robbe-Grillet>,  <Alain Robbe-Grillet; spouse; Catherine Robbe-Grillet> \\
\bottomrule    
\end{tabular}
\end{small}
\vspace{-0.5em}
\caption{Examples of manually labeled relevant knowledge triples for multi-hop questions on the 2Wiki dataset.} 
\label{table:manually_labeled_relevant_triples}
\end{table*}

\subsection{Details and Examples of Manually Labeled Relevant Knowledge Triples}
\label{app:details_examples_relevant_triples}
To quantitatively evaluate the quality of knowledge triples retrieved using our proposed knowledge-driven iterative retrieval framework, we manually label relevant knowledge triples for 100 questions randomly sampled from the 2Wiki dataset. Specifically, for each multi-hop question and its ground-truth relevant documents, we use Llama3 to extract knowledge triples from these relevant documents, and then manually select a subset of knowledge triples that directly support answering the question. 
We provide some examples of the manually curated data in Table~\ref{table:manually_labeled_relevant_triples}.

\begin{figure}[tb]
\begin{center}
\includegraphics[width=0.48\textwidth]{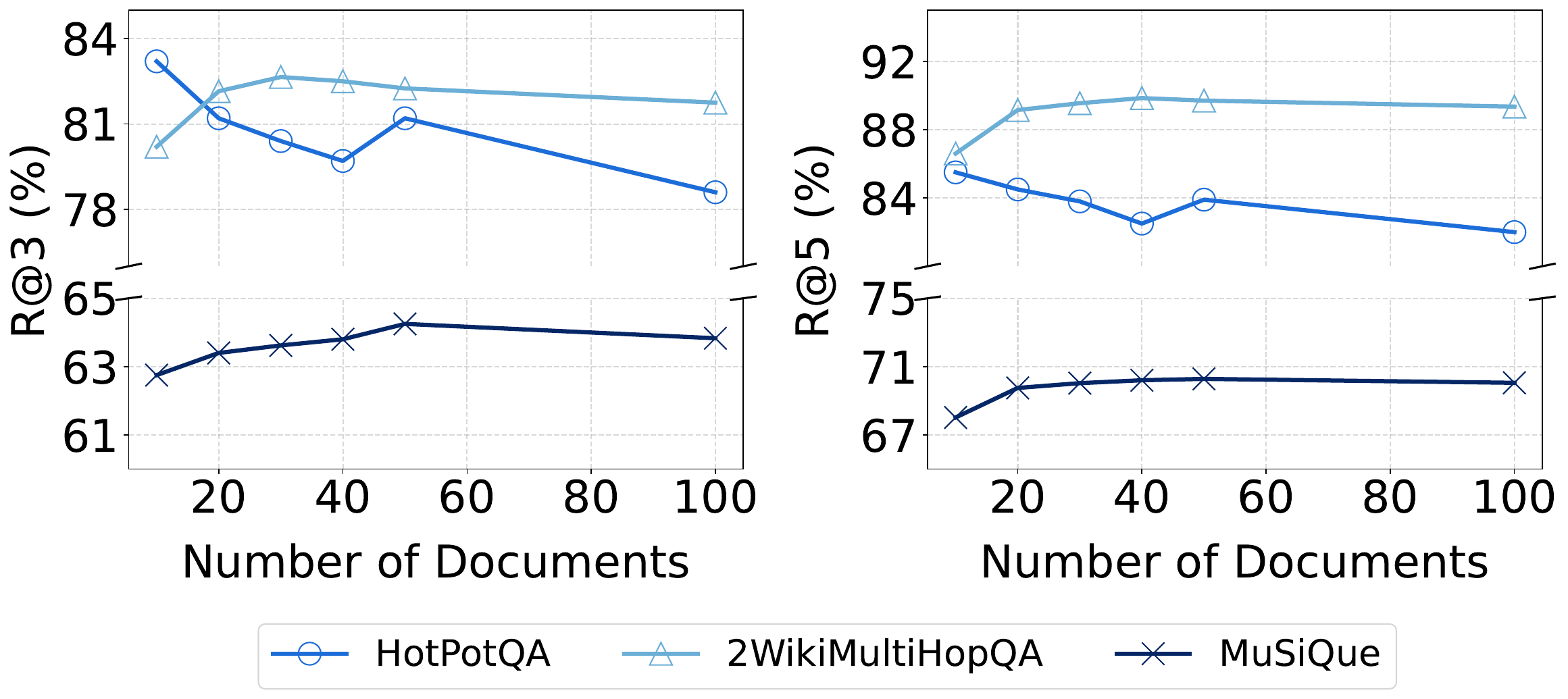}
\end{center}
\vspace{-0.5em}
\caption{Retrieval performance (\%) of \OURS{} under different values of $K_0$ on three multi-hop QA datasets.}
\vspace{-0.5em}
\label{figure:effect_num_documents}
\end{figure}

\subsection{Effect of the Number of Initially Retrieved Documents}
\label{app:effect_of_m}
During the iterative retrieval process of \OURS{}, the Retriever model initially retrieves $K_0$ documents from the corpus, from which relevant knowledge can be extracted. To examine the impact of $K_0$, we vary its value from 10 to 100. Figure~\ref{figure:effect_num_documents} illustrates the retrieval performance of \OURS{} under different values of $K_0$ on the development sets of three multi-hop QA datasets.  
The results indicate that increasing $K_0$ beyond a certain point can degrade performance. This occurs because a larger document pool raises the likelihood of including noisy or irrelevant knowledge triples, making it more challenging for the Reasoning Chain Aligner to accurately identify the triples essential for answering multi-hop questions. 
Therefore, it is crucial to select a proper $K_0$ to achieve superior retrieval performance.

\begin{figure}[tb]
\begin{center}
\includegraphics[width=0.48\textwidth]{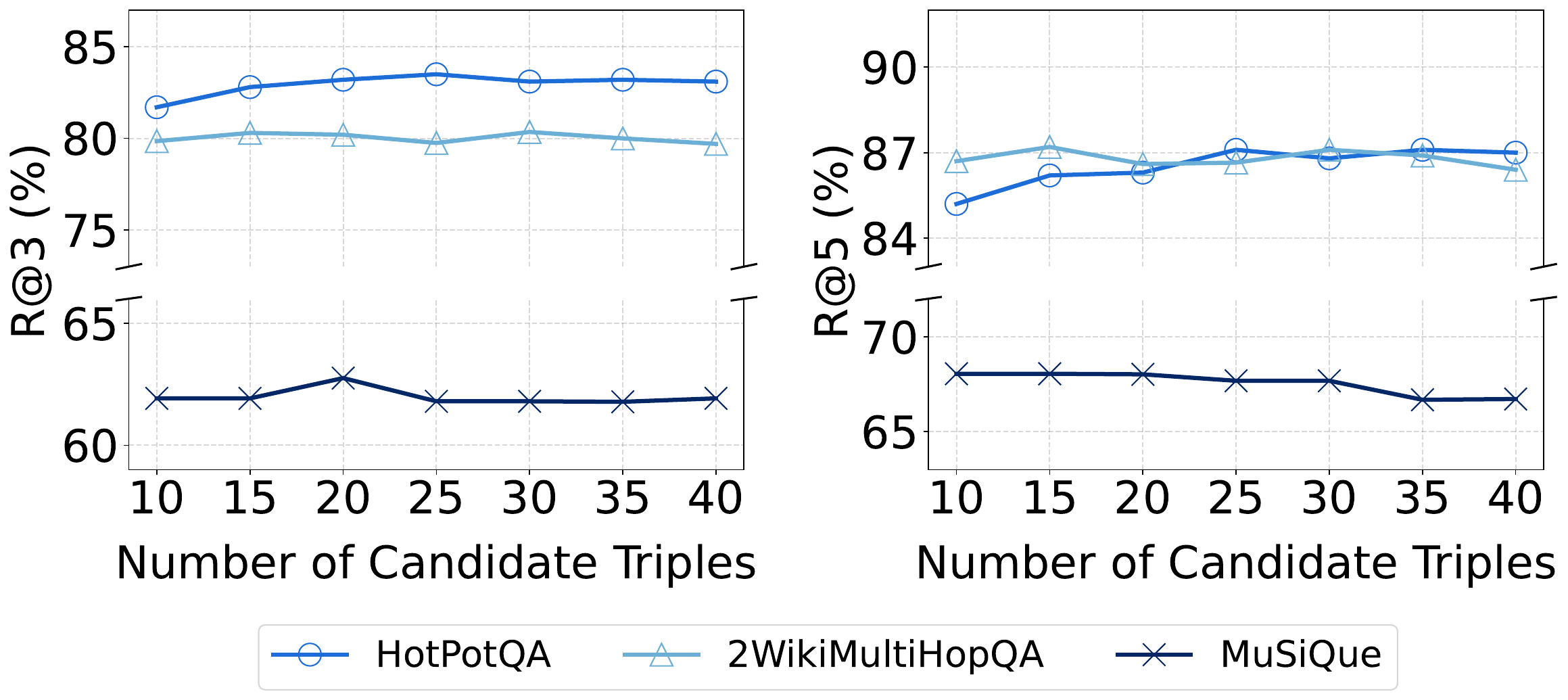}
\end{center}
\vspace{-0.5em}
\caption{Retrieval performance (\%) of \OURS{} under different values of $N$ on three multi-hop QA datasets.}
\label{figure:effect_num_candidate_triples}
\vspace{-0.5em}
\end{figure}

\subsection{Effect of the Number of Candidate Triples} 
\label{app:effect_of_n}
In the iterative retrieval process of \OURS{}, the Reasoning Chain Aligner selects $N$ knowledge triples that are most likely to form a coherent reasoning chain with the existing chain. 
To investigate the effect of $N$, we vary its value from 10 to 40. Figure~\ref{figure:effect_num_candidate_triples} shows the retrieval performance of \OURS{} under different values of $N$ on the development sets of three multi-hop QA datasets. The results indicate that \OURS{} is not sensitive to the value of $N$, as the performance remains relatively stable across different values. This stability can be attributed to the powerful reasoning and contextual understanding abilities of the Reasoning Chain Constructor, which effectively identifies the most useful triple even from a potentially noisy set of candidates triples.

\begin{figure}[tb]
\begin{center}
\includegraphics[width=0.45\textwidth]{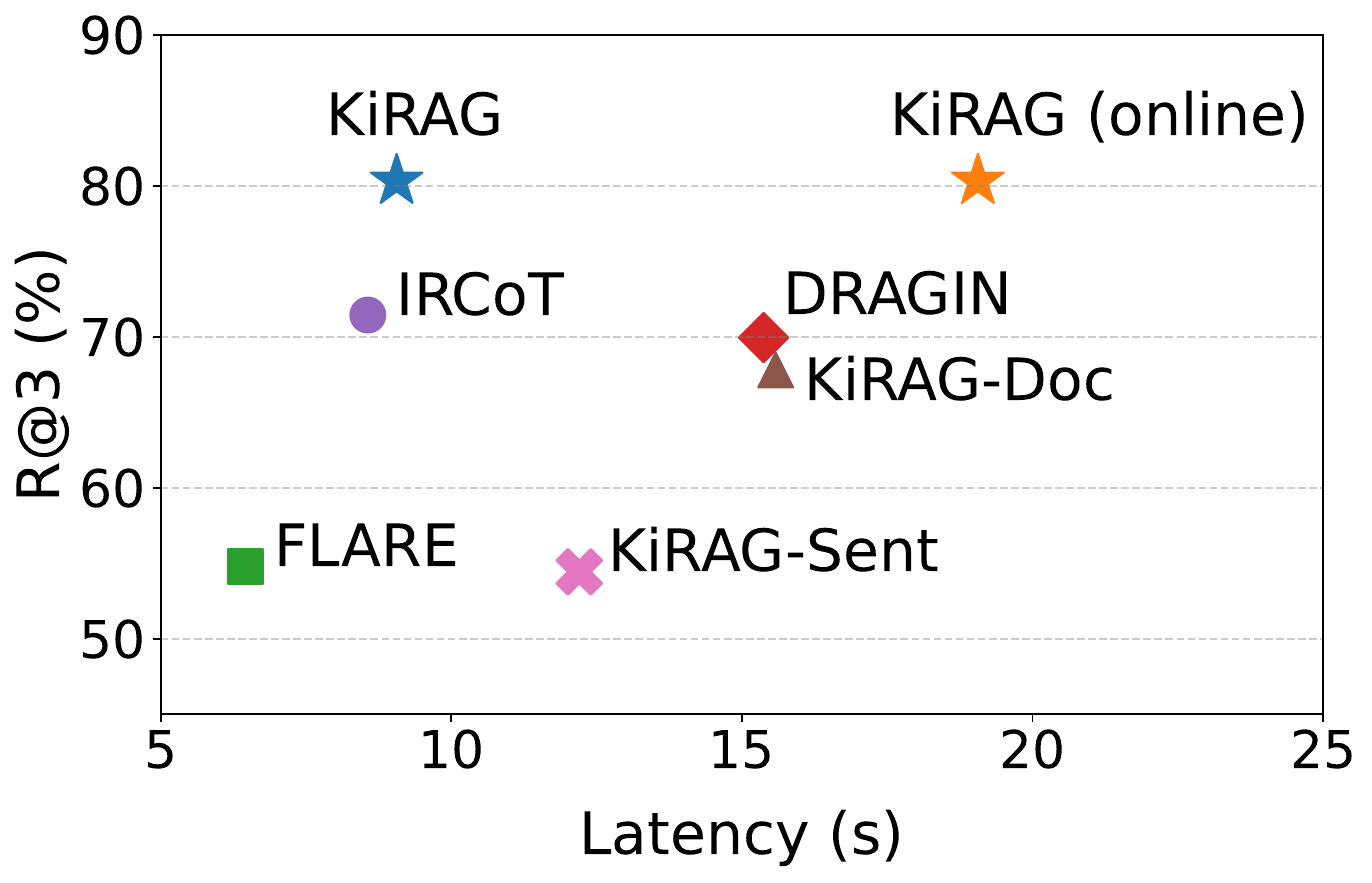}
\end{center}
\vspace{-1.0em}
\caption{Retrieval performance (R$\text{@}$3) v.s. average latency per question for different models on the HotPotQA test set. \OURS{} (online) represents a variant of our approach where knowledge triples extracted dynamically during iterative retrieval, without precomputation.}
\label{figure:efficiency_analysis}
\vspace{-0.5em}
\end{figure}

\subsection{Efficiency Analysis}
\label{app:efficiency_analysis}
We evaluate the efficiency of \OURS{} in comparison to the baseline models. 
Specifically, we conduct experiments on a 3.5 GHZ, 32-cores AMD Ryzen Threadripper Process paired with an NVIDIA A6000 GPU. For fair comparison, both \OURS{} and baselines leverage the same E5 model for document retrieval and the same Llama3 model as the reasoning component. 
It is worth noting that the knowledge triple extraction in our \OURS{} is query-independent and precomputed, which helps to improve efficiency. % To evaluate its impact, 
\jy{To evaluate the impact of precomputing triples,} we introduce a variant: \textit{\OURS{} (online)}, where knowledge triples are dynamically extracted during the iterative retrieval process. 

Figure~\ref{figure:efficiency_analysis} presents the average latency and retrieval performance of different models on the HotPotQA test set, which yields the following findings:
(1) Compared with \OURS{} (online), \OURS{} substantially reduces latency without compromising retrieval performance, highlighting the efficiency benefits of precomputed knowledge triple extraction; 
(2) \OURS{} exhibits latency comparable to IRCoT while achieving significantly better retrieval performance, indicating that our approach effectively enhances retrieval effectiveness without introducing substantial computational overhead. 
(3) \OURS{} achieves a better balance between retrieval effectiveness and efficiency compared to baselines, as evidenced by the relatively lower latency and higher retrieval recall.

\begin{table}[tb]
\centering
\resizebox{0.48\textwidth}{!}{
\begin{tabular}{lcccccc}
\toprule
\multirow{2}{*}{\textbf{Model}} & \multicolumn{2}{c}{\textbf{HotPotQA}} & \multicolumn{2}{c}{\textbf{2Wiki}} & \multicolumn{2}{c}{\textbf{MuSiQue}}\\
\cmidrule(lr){2-3} \cmidrule(lr){4-5} \cmidrule(lr){6-7}
& \textbf{R}$\text{@}$\textbf{3} & \textbf{R}$\text{@}$\textbf{5} & \textbf{R}$\text{@}$\textbf{3} & \textbf{R}$\text{@}$\textbf{5} & \textbf{R}$\text{@}$\textbf{3} & \textbf{R}$\text{@}$\textbf{5} \\
\midrule
\textbf{IRCoT} & {71.44} & {77.57} & {64.30} & {75.56} &45.61 &52.21 \\
\hdashline\noalign{\vskip 0.5ex}
\textbf{\OURS{} (Llama3)} &\textbf{80.32} & \textbf{84.08} & \textbf{77.76} & \textbf{85.32} & {54.53} & {61.16} \\
\textbf{\OURS{} (Mistral)} &74.14 & 79.51 & 74.14 & 82.30 & 49.10 & 56.65 \\
\textbf{\OURS{} (Gemma2)} & 79.66 & 84.03 & 77.04 & 83.59 & \textbf{54.82} & \textbf{62.42} \\
\bottomrule    
\end{tabular}
}
\vspace{-0.5em}
\caption{Retrieval performance (\%) of \OURS{} using different LLM-based Reasoning Chain Constructor.}
\vspace{-0.5em}
\label{table:effects_different_llm_based_constructor}
\end{table}

\subsection{Performance of Using Different LLM-Based Constructor}
\label{app:performance_different_llm}
\OURS{} leverages a frozen LLM as the Reasoning Chain Constructor to maintain the adaptability of our framework. 
Figure~\ref{table:effects_different_llm_based_constructor} presents the retrieval performance of our \OURS{} using different LLM-based Constructor. The results indicate that \OURS{} consistently outperforms IRCoT across different Constructors, indicating the robustness of our approach in improving retrieval performance regardless of the specific LLM used. 

\end{document}